\documentclass[english,12pt, letter]{article}
\usepackage{natbib}
\usepackage{breakcites}
 \bibpunct[, ]{(}{)}{,}{a}{}{,}%
\usepackage{amssymb,amsmath,amsthm,graphicx} %PLAIN Apr 2018

\usepackage{soul}
\usepackage{microtype}
\usepackage{subfigure}
\usepackage{comment,amsfonts,mathtools,bm,booktabs,lmodern, microtype,xcolor,algorithm}
\usepackage{enumerate}
\graphicspath{ {./} {./figures/} }
\usepackage[noend]{algorithmic}
\usepackage{hyperref}
\hypersetup{colorlinks=true,citecolor=red}
\usepackage{geometry}
\usepackage{autobreak}
\usepackage{bbm}
\usepackage[english]{babel}
\newtheorem{lemma}{Lemma}
\newtheorem{theorem}{Theorem}

\newcommand{\exputility}{\mathbb{E}[U(\mathbf{S};\mathbf{u},q)]}

\newcommand{\expected}{\mathbb{E}}

\begin{document}
\title{Thompson Sampling for a Fatigue-aware Online Recommendation System}

\author{Yunjuan Wang \\ Electrical and Computer Engineering \\ University of Illinois at Chicago \\ \url{ywang581@uic.edu} \and Theja Tulabandhula \\ Information and Decision Sciences \\ University of Illinois at Chicago  \\ \url{tt@theja.org} }

\date{Jan 23, 2019}

\maketitle

\begin{abstract}
In this paper we consider an online recommendation setting, where a platform recommends a sequence of items to its users at every time period. The users respond by selecting one of the items recommended or abandon the platform due to fatigue from seeing less useful items. Assuming a parametric stochastic model of user behavior, which captures positional effects of these items as well as the abandoning behavior of users, the platform's goal is to recommend sequences of items that are competitive to the single best sequence of items in hindsight, without knowing the true user model a priori. Naively applying a stochastic bandit algorithm in this setting leads to an exponential dependence on the number of items. We propose a new Thompson sampling based algorithm with expected regret that is polynomial in the number of items in this combinatorial setting, and performs extremely well in practice. 
\end{abstract}

\section{Introduction}
\label{introduction}

%Introduction to the applied problem
In applications such as email newsletters or app notifications, the platform's goal is to carefully tailor items (for instance, items) so as to maximize revenue while maintaining user retention. Both these metrics depend not only on the intrinsic quality of the items themselves, but also on the way they are positioned when the users view them. When the user's precise behavior is not known a priori, the platform may have to learn and maximize revenue simultaneously. In many such platforms, users can be categorized into types (for instance, based on information such as IP, location etc.) and the platform has the ability to interact with multiple users of the same type sequentially and independently. If the items are well aligned with the interests of the users, then the platform benefits from increased sales, its brand gets promoted and may also cause steady user growth. On the other hand, if the items are not interesting to the users, then it may induce fatigue (a state where their perceived value of the platform decreases) leading to user abandonment (for instance, canceled subscriptions or app uninstalls).

%The stylized model in this paper
In this paper, we consider the following setting: the platform needs to learn a sequence of items (from a set of $N$ items) by interacting with its users in rounds. In particular, it wants to maximize its expected utility when compared to the best sequence in hindsight. When a user is presented with a sequence of items, they view it from top-to-bottom and at each position, we can have the following stochastic outcomes:
\begin{enumerate}
\item The user is satisfied with the current item (perhaps clicks the item's link and navigates to a target page). In this situation, the platform gets a reward. 
\item The user is not satisfied with the current item and is willing to look at the next item (for instance, the next notification) if it exists. Note that, it may happen that the user did not select any item and has reached the end of the sequence. In this case, the platform does not get a reward but is also not explicitly penalized.
\item The user has lost interest in the platform (presumably after viewing un-interesting items) and s/he decides to abandons the platform (for instance, by uninstalling the app). In this situation, we ascribe a penalty cost to the platform.
\end{enumerate}

% Vanilla MAB solutions are not enough
One could attempt to model the above problem using the stochastic Multi-armed Bandit (MAB) formalism, where the decision maker selects one arm out of (say $M$) arms in each round, and receives feedback in the form of a reward sampled from a reward distribution. In our setting, each arm would correspond to a sequence of items, and the regret would depend exponentially on the number of items. 

%Problem challenges and our solution
In our setting, the platform can choose both the length of the sequence as well as the order of the items, and this is essentially a combinatorial problem in each round. The recommended sequence of items should balance the penalty of user abandonment versus the upside of user choosing a high revenue item. The probability of a user choosing a high revenue item is not independent of other items in the recommended list. We assume that the aforementioned user behavior has a particular parametric form (detailed in Section~\ref{model}), whose parameters are not known a priori. Our main contribution is the design of a fatigue-aware online recommendation solution, which we call the \emph{Sequential Bandit Online Recommendation System} (SBORS). SBORS, which is based on Thompson sampling, comes with attractive regret guarantees and makes an ordered list of item recommendations to users by carefully exploring their suitability and exploiting learned information based on previous user feedback.

The key contributions of this paper are as follows: First, we design a Thompson sampling (TS) based  algorithm (Section~\ref{algorithm}) for the online fatigue-aware recommendation problem with unknown user preference and abandonment distributions. Second, we formally present SBORS by modifying the above algorithm with posterior approximation and correlated sampling to control exploration-vs-exploitation trade-off. We give detailed analysis of SBORS (Section~\ref{regretanalysis}, supplementary) and prove that the regret upper bound is $C_{1}N^{2}\sqrt{NT\log TR}+C_{2}N\sqrt{T\log TR\cdot \log T}+C_{3}N/R$ (here $C_{1},C_{2}$ and $C_{3}$ are constants, and $R$ is a tunable algorithm parameter that captures exploration-exploitation tradeoff via sampling, see Section \ref{algorithm}). Third, we experiment with our algorithm under several conditions, contrasting it with competing baselines~\cite{aroyo2018reports}, and show that it performs favorably (Section~\ref{experiments}). 
\section{Literature Review}
\label{literature}

\subsection{Multi-armed bandit problem}
The multi-armed bandit problem \cite{lai1985asymptotically,berry1985bandit,sutton2018reinforcement,auer2002finite} is a classic reinforcement learning problem that exemplifies the exploration-exploitation trade-off dilemma. In the traditional version of this problem, the decision maker selects one out of $M$ arms in each round and receives a stochastic reward corresponding to the selected arm. Since each arm has an unknown reward distribution, the goal is to be close the the performance of the best arm (that give the highest expected reward) after multiple rounds. A variant of this MAB problem is cast in the combinatorial setting, where in each round we select an arm that can be viewed as being composed of a set of base elements~\cite{chen2013combinatorial,chen2016combinatorial,wang2017improving}.

There are many approaches to solve the stochastic bandit problem. One of the mainstream methods is the Upper Confidence Bound (UCB) algorithm~ \cite{auer2002using,bubeck2012regret,chen2013combinatorial} (and its many variations). An alternative approach that is different from the UCB family, is the Thompson sampling (TS) approach~\cite{agrawal2012analysis,russo2018tutorial,kaufmann2012thompson}. Extensions of these to contextual settings have also been investigated~\cite{li2010contextual,cheung2017thompson} that allow for richer decision making models and algorithms. While some prior work \cite{wang2018thompson,durand2014thompson} has studied the application of the TS methodology to the stochastic combinatorial multi-armed bandit problem, the combinatorial structure they exploit is not enough to be useful in out setting, or their regret upper bounds or too loose. In our setting, the feasible decisions are sequences of items, which are richer than other objects such as sets.

\subsection{Assortment optimization problem}

For a particular combinatorial problem, namely the assortment optimization problem, \cite{agrawal2017mnl} and \cite{agrawal2017thompson} provide UCB and TS based approaches with attractive regret guarantees. Assortment optimization is the task of choosing a set of items that maximizes the expected revenue assuming a user behavior model (similar to our setting). A particular variant of this problem was initially studied in \cite{rusmevichientong2010dynamic,saure2013optimal} and further discussed by \cite{davis2013assortment,desir2014near,gallego2014constrained,agrawal2017mnl,agrawal2017thompson,agrawal2016near}. Since the number of sets is exponential in the number of items, direct application of a MAB solution turns out to be suboptimal. Similar to \cite{agrawal2017thompson}, we develop a new algorithm for our online recommendation problem (called SBORS) that comes with attractive regret guarantees. The key difference with assortment optimization is that the problem is polynomially solvable in each round whereas in our case the computational problem in each round is NP-hard. We also consider fatigue, which is not present in assortment optimization. Our analysis builds on the machinery developed by \cite{agrawal2017thompson} and uses correlated sampling to control exploration-exploitation trade-off.

\subsection{Sequential choice bandit problem}
The basic form of sequential choice bandit problem, developed by \cite{craswell2008experimental}, is a cascade model where a user views search results displayed by web engine from top to bottom and clicks the first attractive one. \cite{kveton2015cascading} present an online learning version of the cascade model where the platform receives a reward if a user clicks one item, and solve it using a UCB based algorithm. \cite{wangchicheung18} propose a Thompson Sampling based algorithm to minimize regret under the cascade model. Similarly, the setting in \cite{aroyo2018reports} takes the probability of abandoning the platform into consideration, which can be regarded as an extension of the basic cascade model. 

In particular, \cite{aroyo2018reports} use an UCB based approach to recommend a sequence of messages to users under the same user behavior model studied in this work. Their key novelties include showing that the combinatorial problem is linear time solvable and providing a tight regret upper bound (O$(N\sqrt{T \log T})$), where $N$ is the number of messages and $T$ is the horizon (total number of rounds). In general, the combinatorial problem is NP-hard (for instance, when we have capacity constraints, in contrast to the assortment optimization problem where it is still polynomial time). Further, we show in this paper that a TS based approach outperforms their algorithm empirically over a wide range of problem instances (although we get a slightly worse upper bound of O($N^2\sqrt{NT\log T}$)). Hence, our contribution complements their results and allows for a complete understanding of the fatigue-aware online recommendation problem.
\section{Model}
\label{model}
Our setting is similar to that of \cite{aroyo2018reports}. Consider a platform containing $N$ different items indexed by $i$. let its corresponding revenue be $r_{i}$ if selected. User's intrinsic preference for an item $i$ is denoted by $u_{i}$. After viewing each item from a recommended list, the user has a probability $p$ of abandoning the platform, and the occurrence of this event causes the platform to incur a penalty cost $c$. Note that $r_{i}, u_{i}, q, c \in [0,1]$. We represent the sequence of items at time/round $t$ as $\mathbf{S}^{t}=(S_{1}^{t}, S_{2}^{t},..., S_{m}^{t})$, where $S_{i}^{t}$ denotes the $i^{th}$ item, and $m$ represents the length of the sequence.

After the user at time $t$ sees item $i$, s/he has three options based on behavior parameters $\mathbf{u}$ and $p$: 
(1) The user  is satisfied with the item $i$, then no further items are presented to the user. In this situation, the platform earns revenue $r_{i}$. 
(2) The user is not satisfied with item $i$ and decides to see the following item $i+1$ in the sequence of items.%, it it exists.
When the sequence runs out, the user exits the platform. In this situation, the platform will neither earn a reward nor pay a penalty cost. 
(3) The user is unsatisfied with the platform altogether after looking at some items, and s/he abandons the platform. In this situation, the platform incurs a penalty $c$.

The behavior parameters $\mathbf{u}$ and $p$ parameterize the following distributions. Consider a random variable $W^{t}$ following a distribution $F_{W}$. $W^{t}$ measures the $t^{th}$ user's patience, capturing the number of unsatisfied items the user sees without abandoning the platform. In particular, $F_{W}$ is a geometric distribution with parameter $p$. Let $q=1-p$. Then $q_{k}=q^{k-1}(1-q)$ denotes the probability that a user abandons the platform after receiving $k^{th}$ unsatisfying item. Further, let $\widetilde{F}_{W}(k)=P(W>k)=1-P(W\leq k)=q^{k}$ denote the probability that a user does not abandon the platform after receiving the $k^{th}$ unsatisfying item. The probability of each item $i$ being selected is $u_{i}$, which is only determined by its content. The probability of each item $i$ being selected when it belongs to the sequence of items $\mathbf{S}$ (dropping the superscript $t$ for simplicity) is denoted as $p_{i}(\mathbf{S})$. $p_{i}(\mathbf{S})$ not only depends on the item's intrinsic value to the user, but also depends on its position and the other items shown before it. The probability of total abandonment is denoted as $p_{a}(\mathbf{S})$, and represents the sum of the probabilities that the platform is abandoned after receiving $k$ unsatisfying items. In summary,
\begin{eqnarray*}
\begin{aligned}
p_{i}(\mathbf{S})=
&\left\{\begin{array}{ccc}
  u_{i} &\textrm{if }i \in S_{1},\\
  \widetilde{F}_{W}(l-1)\prod_{k=1}^{l-1}(1-u_{I(k)})u_{i} & \textrm{if }i\in S_{l},l\geq 2,\\
  0 &\textrm{if }i\notin S.\\
\end{array}\right.
\end{aligned}
\end{eqnarray*}

And $p_{a}(\mathbf{S})=\sum_{k=1}^{m}q_{k}\prod_{j=1}^{k}(1-u_{I(j)})$, where $I(k)$ means that in the sequence of items $\mathbf{S}$, the $k^{th}$ items is $i$, i.e. $S_{k}=\{i\}$. We denote $U(\mathbf{S},\mathbf{u},q)$ as the total utility (payoff) that the platform receives from a given sequence of items $\mathbf{S}$. The goal is to find the optimal sequence of items that can optimize the expected utility $\exputility=\sum_{i\in \mathbf{S}}p_{i}(\mathbf{S})r_{i}-cp_{a}(\mathbf{S})$:
\begin{eqnarray}
\begin{aligned}
&\underset{\mathbf{S}}{\max}\textrm{\ \ \ } \exputility\\
&\textrm{ s.t. }\quad S_{i}\cap S_{j}=\emptyset, \forall i\neq j.
\end{aligned}
\label{eqn:optim}
\end{eqnarray}
The constraint above specifies that all the items contained in the sequence are distinct. We denote the optimal sequence of items for a given $\mathbf{u},q$ pair using $\mathbf{S}^{*}=\underset{\mathbf{S}}{\arg\max}\exputility$. If it is not unique, ties are broken arbitrarily. 

\section{Algorithm}
\label{algorithm}

A key aspect of our online recommendation algorithm SBORS (which is based on TS) will be that it solves the optimization problem (\ref{eqn:optim}) in each round. We first discuss the complexity of this problem and a precursor to SBORS.

\subsection{The combinatorial problem}\label{combinatorial}

To start, we first define a binary decision variable $f_{i,k}$ to represent the choice of positioning item $i$ at location $k$  in a sequence of items. These variables are constrained as follows: First, since each item can be chosen at most once, it corresponds to the constraint $\sum_{k}f_{i,k}\leq 1, \forall i \leq N$. Second, one position can only place one item. Thus we have $\sum_{i}f_{i,k}\leq 1, \forall k \leq N$. These constraints are not enough to represent sequences without \emph{gaps} (no item in a position followed by an item in the next position), so we use a proxy variable $g(k)$ which denotes the actual position of an item if $f_{i,k}$ is $1$. The optimization problem can be written as:
\begin{eqnarray*}
\begin{aligned}
\underset{\mathbf{F}}{\max}
&\sum_{k=1}^{N}\sum_{i=1}^{N}\bigg[
\big(f_{i,1}u_{i}+(1-f_{i,1})(q^{g(k)-1}\prod_{j=1}^{k-1}(1-\sum_{\ell=1}^{\ell=N}u_{\ell}f_{\ell,j})u_{i})\big)r_{i}\\
&\quad -cq^{g(k)-1}(1-q)\prod_{j=1}^{k}(1-\sum_{\ell=1}^{N}u_{\ell}f_{\ell,j})
\bigg]f_{i,k}
\\
s.t.\textrm{\ \ }
&\sum_{k=1}^{N}f_{i,k}\leq 1, \forall i \in [N]\\
& \sum_{i=1}^{N}f_{i,k}\leq 1, \forall k \in [N]\\
& g(1)= 1, \textrm{ and } f_{i,k}\in\{0,1\}, \forall i,k \in [N],
\end{aligned}
\end{eqnarray*}
where $g(k)=\sum_{j=1}^{k}\sum_{i=1}^{N}f_{i,j}$. Additional constraints on the decision variables (for instance, motivated by business rules such as an upper bound on the sequence length or some diversity requirement on the sequence) can render the problem NP-hard. Without additional constraints however, the problem is linear time solvable, as shown below.

\begin{theorem}\cite{aroyo2018reports}
\label{theorder}
For item $i\in \{1,...,N\}$, define its score as $\theta_{i}:=\frac{r_{i}u_{i}-cp(1-u_{i})}{1-q(1-u_{i})}$. Without loss of generality, assume items are sorted in the decreasing order of their scores, i.e., $\theta_{1}\geq\theta_{2}\geq\cdots\geq\theta_{N}$. Then the optimal sequence of items is $\mathbf{S}^{*}=(1,2,...,m)$, where $m=\max\{i:r_{i}u_{i}-cp(1-u_{i})>0\}$.
\end{theorem}

If the feasible set of solutions in enumerable (for instance it is polynomial in $N$), then an alternative strategy is to perform a sub-linear (in the number of feasible solutions) time search using Locality Sensitive Hashing~\cite{sinha2017optimizing}. Our algorithm SBORS relies on an oracle solving the above problem for a given input of $\mathbf{u},q$ pair. For the remainder, we will assume that such an oracle exists and focus on the exploration-exploitation trade-off.

\subsection{Precursor to SBORS: independent Beta priors}

We first describe an algorithm that captures the TS approach. Unfortunately, a direct analysis of this version is difficult, so we modify it suitably to design our proposed algorithm SBORS in Section \ref{secsbors} later on. TS involves maintaining a posterior on the unknown parameters, which is updated every time new feedback is observed. In the beginning of every round, the parameters are sampled from the current posterior distribution, and the algorithm chooses the best sequence of items based on these sampled parameters.

Denote $c_{i}(t)$ as the total number of users selecting item $i$, and $f_{i}(t)$ as the total number of users observing item $i$ without selection. Let $T_{i}(t)=c_{i}(t)+f_{i}(t)$. Denote $n_{a}(t)$ as the number of users who abandon the platform by time $t$, $n_{e}(t)$ as the number of times that users do not select an item and do not abandonment by time $t$. Let $N_{q}(t)=n_{e}(t)+n_{a}(t)$. Let $I(\cdot)$ denote the index function such that $I(k)=i$ if and only if $S_{k}={i}$. As shown in \cite{aroyo2018reports} (Lemma 5), we can get unbiased estimates of the true parameters as follows: 

\begin{lemma}
\label{unbiasedes}
Unbiased estimates: $\hat{u}_{i}(t)=\frac{c_{i}(t)}{T_{i}(t)}$ is an unbiased estimator for $u_{i}$ and $\hat{q}_{i}(t)=\frac{n_{e}(t)}{N_{q}(t)}$ is an unbiased estimator for q.
\end{lemma}

In this version of the algorithm, we maintain a Beta posterior distribution for the selection parameter $u_i$ and the abandonment distribution parameter $q$, which we update as we observe the user's feedback to our current recommended list. 
At the initial state, $u_{i}$ and $q$ are unknown to the platform, $r_{i}$ and $c$ are known to the platform.
For a user arriving at time $t$, we calculate the current optimal sequence of items based on samples $\mathbf{u}'(t)$ and $q'(t)$. When the sequence of items is shown, the user has three options: (1) select one item and leave the interface; (2) see all the items without selection and abandonment; or (3) abandon the platform. After each round, we update the parameters of the relevant Beta distributions.

\begin{algorithm}[ht!]
\caption{TS-based algorithm (precursor to SBORS)}\label{preSBORS}
\begin{algorithmic}
    \STATE {\bfseries Initialization:}
    Set $c_{i}(t)=f_{i}(t)=1$ for all $i\in X$; $n_{e}(t)=n_{a}(t)=1$; and $t=1$;

        \WHILE{$t\leq T$}
        \item[]{(a) $Posterior\ sampling$:}
        %\begin{itemize}
        
        {For each item $i=1,...,N$, sample $u_{i}'(t)$ and $q'(t)$}

        {$u_{i}'(t)\sim Beta(c_{i}(t),f_{i}(t))$}

        {$q'(t)\sim Beta(n_{e}(t),n_{a}(t))$}
        %\end{itemize}
        \item[]{(b) $Sequence\ selection$:}
        
        Compute $\mathbf{S}^{t}=\underset{\mathbf{S}}{\arg\max} \expected[U(\mathbf{S};\mathbf{u}'(t),q'(t))]$;

        Observe feedback upon seeing the $k_{t} \leq |\mathbf{S}^{t}|$ items;

        \item[]{(c) $Posterior\ update$: }
        \FOR {$j=1,\cdots,k_{t}$}

        \item[] Update 
            \begin{eqnarray*}
            \begin{aligned}
            (c_{I(j)}(t),f_{I(j)}(t),n_{e}(t),n_{a}(t))=\left\{\begin{array}{c}
            (c_{I(j)}(t)+1,f_{I(j)}(t),n_{e}(t),n_{a}(t))\\ 
            \quad \textrm{ if select and leave} \\
            (c_{I(j)}(t),f_{I(j)}(t)+1,n_{e}(t)+1,n_{a}(t)) \\
            \quad \textrm{ if not select and not abandon}\\
            (c_{I(j)}(t),f_{I(j)}(t)+1,n_{e}(t),n_{a}(t)+1) \\
            \quad \textrm{ if not select and abandon}\\
            \end{array}\right.
            \end{aligned}
            \end{eqnarray*}
        \ENDFOR
        \item[]{
        $c_{i}(t+1)=c_{i}(t)$,
        $f_{i}(t+1)=f_{i}(t)$ for all $i \in [N]$
        
        $n_{e}(t+1)=n_{e}(t)$,
        $n_{a}(t+1)=n_{a}(t)$
        
        $t=t+1$}
        \ENDWHILE
\end{algorithmic}
\end{algorithm}

\subsection{SBORS: Sequential Bandit for Online Recommendation System}
\label{secsbors}

Motivated by \cite{agrawal2017thompson}, we modify Algorithm \ref{preSBORS} by: (a) introducing a posterior approximation by Gaussians, and (b) performing correlated sampling (which boosts variance boosting and allows for a finer exploration-exploitation trade-off). 

\textbf{Posterior approximation:} We approximate the posteriors for $u_i$, $q$ by Gaussian distributions with approximately the same mean and variance as the original Beta distributions. In particular, let
\begin{eqnarray}
\begin{aligned}
\label{hatumeansigma}
\hat{u}_{i}(t)&=\frac{c_{i}(t)}{c_{i}(t)+f_{i}(t)}=\frac{c_{i}(t)}{T_{i}(t)},\\
\hat{\sigma}_{u_{i}}(t)&=\sqrt{\frac{\alpha\hat{u}_{i}(t)(1-\hat{u}_{i}(t))}{T_{i}(t)+1}}+\sqrt{\frac{\beta}{T_{i}(t)}},
\end{aligned}
\end{eqnarray}
\begin{eqnarray}
\begin{aligned}
\label{hatqmeansigma}
\hat{q}(t)&=\frac{n_{e}(t)}{n_{e}(t)+n_{a}(t)}=\frac{n_{e}(t)}{N_{q}(t)}, \textrm{ and}\\
\hat{\sigma}_{q}(t)&=\sqrt{\frac{\alpha\hat{q}(t)(1-\hat{q}(t))}{N_{q}(t)+1}}+\sqrt{\frac{\beta}{N_{q}(t)}},
\end{aligned}
\end{eqnarray}
where $\alpha > 0, \beta \geq 2$ are constants, be the means and standard deviations of the approximating Gaussians.

\textbf{Controlling exploration via correlated sampling:} Instead of sampling $\mathbf{u}'$ and $q'$ independently, we correlate them by using a common standard Gaussian sample and transforming it. That is, in the beginning of a round $t$, we generate a sample from the standard Gaussian $\theta \sim N(0,1)$, and the posterior sample for item $i$ is computed as $\hat{u}_{i}(t)+\theta\hat{\sigma}_{u_{i}}(t)$, while the posterior sample for abandonment is computed as $\hat{q}(t)+\theta\hat{\sigma}_{q}(t)$. This allows us to generate sample parameters for $i=1,\cdots,N$ that are highly likely to be either simultaneously high or simultaneously low. As a consequence, the parameters corresponding to items in the ground truth $\mathbf{S}^*$, will also be simultaneously high/low. Because correlated sampling decreases the joint variance of the sample, we can counteract by generating multiple Gaussian samples. In particular, we generate $R$ independent samples from the standard Gaussian, $\theta^{(j)}\sim N(0,1)$, $j \in [R]$, and the $j^{th}$ sample of parameters is generated as:
\begin{eqnarray*}
\begin{aligned}
u_{i}'^{(j)}&=\hat{u}_{i}+\theta^{(j)}\hat{\sigma}_{u_{i}}, \quad \textrm{ and } q^{'(j)} =\hat{q}+\theta^{(j)}\hat{\sigma}_{q}.
\end{aligned}
\end{eqnarray*}
We then use the highest valued samples by simply taking the maximums $
u_{i}'(t)=\underset{j=1,\cdots,R}{\max}u_{i}'^{(j)}(t),$ and $q'(t)=\underset{j=1,\cdots,R}{\max}q'^{(j)}(t)$. These are then used in the optimization problem to get $\mathbf{S}^t\allowbreak=\underset{\mathbf{S}}{\arg\max}\expected[U(\mathbf{S}^{t};\mathbf{u}'(t),q'(t))]$.

Algorithm \ref{preSBORS} samples from the posterior distribution of $\mathbf{u}$ and $q$ independently in each round, which makes the probability of being optimistic (i.e. the optimal sequence of items $\mathbf{S}^{*}$ has at least as much reward on the sampled parameters as on the true parameters) exponentially small. We use correlation sampling to ensure that the probability of an optimistic round is high enough. A detailed explanation is provided in Section \ref{regretanalysis}.

\begin{algorithm}[ht!]
\caption{SBORS algorithm}\label{sbors}
\begin{algorithmic}
    \STATE {\bfseries Initialization:}
   Set $c_{i}(t)=f_{i}(t)=1$ for all $i\in X$; $n_{e}(t)=n_{a}(t)=1$; $t=1$;
    
        \WHILE{$t\leq T$}

         \STATE Update $\hat{u}_{i}(t),\hat{q}(t),\hat{\sigma}_{u_{i}}(t),\hat{\sigma}_{q}(t)$ from (\ref{hatumeansigma}) and (\ref{hatqmeansigma});\\

        \STATE{(a) $Correlated\ sampling$:}

        \FOR {$j=1,...,R$}

        \STATE Get $\theta^{(j)} \sim N(0,1)$ and compute 
        $u_{i}'^{(j)}(t)$,$q^{'(j)}(t)$    
        
        \ENDFOR       
        
        For each $i\leq N$, compute $u_{i}'(t)=\underset{j=1,\cdots,R}{\max}u_{i}'^{(j)}(t)$ and
        $q'(t)=\underset{j=1,\cdots,R}{\max}q'^{(j)}(t)$.

        \STATE{(b) $Sequence\ selection$: }
        Same as step (b) of Algo. \ref{preSBORS}.

        \STATE{(c) $Posterior\ update$: }
        Same as step (c) of Algo. \ref{preSBORS}.
        \ENDWHILE
\end{algorithmic}
\end{algorithm}

\section{Regret Analysis for SBORS}
\label{regretanalysis}

Our main result is the following:
\begin{theorem}
\label{thereg} (\textbf{Main Result})
Over  $T$ rounds, the regret of SBORS (Algorithm \ref{sbors}) is bounded as:
\begin{eqnarray*}
\begin{aligned}
Reg(T;\mathbf{u},q)\leq C_{1}N^{2}\sqrt{NT\log TR}+C_{2}N\sqrt{T\log TR\cdot \log T}+\frac{C_{3}N}{R},
\end{aligned}
\end{eqnarray*}
where $C_{1},C_{2}$ and $C_{3}$ are constants and $R$ is an algorithm parameter.
\end{theorem}

\textbf{Proof Sketch:} We provide a proof sketch below and refer the reader to the supplementary for a more detailed treatment. The pseudo-regret can be expressed as:
\begin{eqnarray*}
Reg(T;\mathbf{u},q)=\expected\left[\sum_{t=1}^{T}\expected[U(\mathbf{S}^{*};\mathbf{u},q)]-\expected[U(\mathbf{S}^{t};\mathbf{u},q)]\right],
\end{eqnarray*}
where $\mathbf{\mathbf{S}^{*}}$ is the optimal sequence when $\mathbf{u}$ and $q$ are known to the platform, while $\mathbf{S}^{t}$ is the sequence offered to the user arriving at time $t$. Adding and subtracting $\sum_{t=1}^{T}\expected[U(\mathbf{S}^{t},\mathbf{u}'(t),q'(t))]$, we can rewrite the regret as $Reg(T;\mathbf{u},q)=Reg_{1}(T,\mathbf{u},q)+Reg_{2}(T,\mathbf{u},q)$ where:
$Reg_{1}(T,\mathbf{u},q)\\
=\expected\left[\sum_{t=1}^{T}\expected[U(\mathbf{S}^{*};\mathbf{u},q)]-\expected[U(\mathbf{S}^{t};\mathbf{u}'(t),q'(t))]\right]$, and
$Reg_{2}(T,\mathbf{u},q)\\
=\expected\left[\sum_{t=1}^{T}\expected[U(\mathbf{S}^{t};\mathbf{u}'(t),q'(t))]-\expected[U(\mathbf{S}^{t};\mathbf{u},q)]\right].$

We say that a round $t$ is optimistic if the optimal sequence of items $\mathbf{S}^{*}$ has at least as much reward on the sampled parameters as on the true parameters, i.e. $\expected[U(\mathbf{S}^{*};\mathbf{u}'(t),q'(t))]\geq \expected[U(\mathbf{S}^{*};\mathbf{u},q)]$.

The first term $Reg_{1}(T,\mathbf{u},q)$ is the difference between the optimal reward given the true parameters $\mathbf{u}$, $q$, and the optimal reward of the sampled sequence of items $\mathbf{S}^{t}$ with respect to the sampled parameters $\mathbf{u}'$, $q'$. Thus this term would contribute \emph{no regret if the round was optimistic}, as defined above. So, we are left to consider only ``non-optimistic'' rounds, which we will show they are not too many in number. Thus, we first prove that at least one of our $R$ samples is optimistic with high probability. Then, we also bound the instantaneous regret of any ``non-optimistic'' round by relating it to the closest optimistic round before it.

The second term $Reg_{2}(T,\mathbf{u},q)$ is the difference in the reward of the offer sequence of items $\mathbf{S}^{t}$ when evaluated on sampled parameters and the true parameters, which can be bounded by the concentration properties of our posterior distributions. The idea is that the expected reward corresponding to the sampled parameters will be close to that on the true parameters. Before elaborating further on the proof details, we first highlight some key lemmas involved in proving Theorem \ref{thereg} below.

\textbf{Key Lemmas:} To analyze the regret, we first provide the concentration results for the relevant quantities. To be specific, the posterior distributions concentrate around their means, which in turn concentrate around the true parameters.

\begin{lemma}
\label{lm}
(Concentration bound) For all $i=1,\cdots,N$, for any $\alpha,\beta,\rho\geq 0$, and $t\in \{1,2,\cdots,T\}$, we have

\begin{eqnarray*}
\begin{aligned}
 P\left(|\hat{u}_{i}(t)-u_{i}|\geq\sqrt{\frac{\alpha\hat{u}_{i}(t)(1-\hat{u}_{i}(t))\log\rho}{T_{i}(t)+1}}+
\sqrt{\frac{\beta\log \rho}{T_{i}(t)}}\right)&\leq \frac{2}{\rho^{2\beta}},\\
P\left(|\hat{q}(t)-q|\geq \sqrt{\frac{\alpha\hat{q}(t)(1-\hat{q}(t))\log \rho}{N_{q}(t)+1}}+\sqrt{\frac{\beta\log \rho}{N_{q}(t)}}\right)&\leq \frac{2}{\rho^{2\beta}}.
\end{aligned}
\end{eqnarray*}
\end{lemma}

\begin{lemma}
\label{anticon}
For any $t\leq T$ and $i \in \{1,\cdots,N\}$, we have for any $r>1$,
\begin{eqnarray*}
&P(|u_{i}'(t)-\hat{u}_{i}(t)|>4\hat{\sigma}_{u_{i}}(t)\sqrt{\log rR})\leq \frac{1}{r^{8}R^{7}}\textrm{ , and}\\
&P(|q'(t)-\hat{q}(t)|>4\hat{\sigma}_{q}(t)\sqrt{\log rR})\leq \frac{1}{r^{8}R^{7}}\textrm{ ,}
\end{eqnarray*}
where $\hat{\sigma}_{u_{i}}(t)$, $\hat{\sigma}_{q}(t)$, $R$, $u_{i}'(t)$, $q'(t)$, $\hat{u}$, $\hat{q}$ are defined in Section \ref{model}.
\end{lemma}

Next we establish two important properties of the optimal expected payoff. The first property is referred to as restricted monotonicity. Simply put, with the optimal sequence of items $\mathbf{S}_{v}^{*}$ determined under some parameters $\mathbf{v}$ and $q_{v}$, its expected payoff is no larger than the payoff under the same sequence of items $\mathbf{S}_{v}^{*}$ when preference parameter $\mathbf{w}$ and the abandonment parameter $q_{w}$ are element-wise larger than $\mathbf{v}$ and $q_{v}$. The second property is a \emph{Lipschitz} style bound on the deviation of the expected payoff with change in the parameters $\mathbf{v}$ and $q_{v}$. To be specific, the difference between the two expected payoffs is bounded by a linear sum of the items' preference and abandonment parameters.

\begin{lemma}
\label{lip}
Suppose $\mathbf{S}_{v}^{*}$ is an optimal sequence of items given $\mathbf{v}$ and $q_{v}$. That is, $\mathbf{S}^{*}_{v}\in\arg\max \expected[U(\mathbf{S},\mathbf{v},q_{v})].$

Then for any $\mathbf{v},\mathbf{w}\in [0,1]^{N}$, $q_{v}, q_{w}\in[0,1]$, we have 

1. (Restricted Monotonicity)
If $v_{i}\leq w_{i}$ for all $i \in [N]$, and $q_{v}\leq q_{w}$, then
$\expected[U(\mathbf{S}_{v}^{*};\mathbf{w},q_{w})]\geq \expected[U(\mathbf{S}_{v}^{*};\mathbf{v},q_{v})].$

2. (Lipschitz)
\begin{eqnarray*}
|\expected[U(\mathbf{S}_{v}^{*};\mathbf{v},q_{v})]-\expected[U(\mathbf{S}_{v}^{*};\mathbf{w},q_{w})]|
\leq\sum_{i \in \mathbf{S}_{v}^{*}}\left(2|v_{i}-w_{i}|+(N+1)|q_{v}-q_{w}|\right).
\end{eqnarray*}
\end{lemma}

From Lemma \ref{lm}, \ref{anticon} and \ref{lip}, we can prove that the difference between the expected payoff of the offered sequence $\mathbf{S}^{t}$ corresponding to the sampled parameters and the true parameters becomes smaller as time increases.

\begin{lemma}
\label{disexp}
For any round $t\leq T$, we have
\begin{eqnarray*}
\begin{aligned}
\expected\bigg\{\expected[U(\mathbf{S}^{t};\mathbf{u}'(t),q'(t))]-\expected[U(\mathbf{S}^{t};\mathbf{u},q)]\bigg\}\leq \expected\bigg[C_{1}'\sum_{i\in \mathbf{S}^{t}}\sqrt{\frac{\log TR}{T_{i}(t)}}+C_{2}'(N+1)\sqrt{\frac{\log TR}{N_{q}(t)}}\bigg],
\end{aligned}
\end{eqnarray*}
where $C_{1}'$ and $C_{2}'$ are universal constants. 
\end{lemma}

We will now discuss how these lemmas can be put together to bound $Reg_{1}(T,\mathbf{u},q)$ and $Reg_{2}(T,\mathbf{u},q)$. 

\noindent\textbf{Bounding the first term $Reg_{1}(T,\mathbf{u},q)$}: Since $\mathbf{S}^{t}$ is an optimal sequence of items for the sampled parameters, we have $\expected[U(\mathbf{S}^{t};\mathbf{u}'(t), q'(t))]\geq \expected[U(\mathbf{S}^{*};\mathbf{u}, q)]$ if round $t$ is optimistic. This suggests that as the number of optimistic round increases, the term $Reg_{1}(T,\mathbf{u},q)$ decreases.

Next, we prove that there are only a limited number of non-optimistic rounds (\emph{this is a key step}). Using a tail bound for the Gaussian distribution, we can control the probability mass associated with the event that a sampled parameter $u_{i}'^{(j)}(t)$ for any item $i$ will exceed the posterior mean by a few standard deviations. Since our Gaussian posterior's mean is equal to the unbiased estimate $\hat{u}_{i}$, and its standard deviation is close to the expected deviation of estimate $\hat{u}_{i}$ from the true parameter $u_{i}$, we can conclude that any sampled parameter $u_{i}'^{(j)}(t)$ will be optimistic with at least a constant probability, i.e., $u_{i}'^{(j)}(t)\geq u_{i}$.
The same reasoning also holds for $q'^{(j)}(t)$. However, for an optimistic round, sampled parameters for all items in $\mathbf{S}^{*}$ needs to be optimistic. This is where the correlated sampling aspect of SBORS is crucially utilized. Using the dependence structure between samples for items in $\mathbf{S}^{*}$, and the variance boosting provided by the sampling of $R$ independent copies, we prove an upper bound of roughly $O(1/R)$ on the number of consecutive rounds between two optimistic rounds. Lemma \ref{eper30n} formalizes this intuition.
\begin{lemma}
\label{eper30n}
(Spacing of optimistic rounds) For any $p\in [1,2]$, we have
\begin{eqnarray*}
\expected^{1/p}\big[|\varepsilon^{An}(\tau)|^{p}\big]\leq \frac{e^{12}}{R}+(C_{3}'N)^{1/p}+C_{4}'^{1/p}
\end{eqnarray*}
where $C_{3}'$ and $C_{4}'$ are constants. $\varepsilon^{An}(\tau)$ is defined as the group of rounds after an optimistic round $\tau$ and before the next consecutive optimistic round. A formal definition of optimistic round is in Section \ref{regretanalysis}.
\end{lemma}

Next, We bound the individual contribution of any ``non-optimistic'' round $t$ by relating it to the closest optimistic round $\tau$ before it. By the definition of an optimistic round,
\begin{eqnarray*}
\expected[U(\mathbf{S}^{*};\mathbf{u},q)]-\expected[U(\mathbf{S}^{t};\mathbf{u}'(t),q'(t))]
\leq \expected[U(\mathbf{S}^{\tau};\mathbf{u}(\tau),q(\tau))]-\expected[U(\mathbf{S}^{t};\mathbf{u}'(t),q'(t))],
\end{eqnarray*}
and by the choice of $\mathbf{S}_{t}$ we get:
\begin{eqnarray*}
\begin{aligned}
\expected[U(\mathbf{S}^{\tau};\mathbf{u}(\tau),q(\tau))]-\expected[U(\mathbf{S}^{t};\mathbf{u}'(t),q'(t))]
\leq \expected[U(\mathbf{S}^{\tau};\mathbf{u}(\tau),q(\tau))]-\expected[U(\mathbf{S}^{\tau};\mathbf{u}'(t),q'(t))].
\end{aligned}
\end{eqnarray*}

What remains to be shown is a bound on the difference in the expected payoff of  $\mathbf{S}^{\tau}$ for $\mathbf{u}(\tau),q(\tau)$ and for $\mathbf{u}'(t),q'(t)$. Over time, as the posterior distributions concentrate around their means, which in turn concentrate around the true parameters, we can show that this difference becomes smaller. As a result, $Reg_{1}$ can be bounded as: $Reg_{1}(T,\mathbf{u},q)\leq O(N\sqrt{T\log TR \log T})+O(N/R).$

\noindent\textbf{Bounding the second term $Reg_{2}(T,\mathbf{u},q)$}: Similar to the discussion above, using the Lipschitz property (Lemma~\ref{lip}) and Lemma \ref{disexp}, this term can be bounded as: $ Reg_{2}(T,\mathbf{u},q)\leq O(N^{2}\sqrt{NT\log TR}).$ Overall, the above analysis on $Reg_{1}$ and $Reg_{2}$ implies the following bound on the overall regret:
\begin{eqnarray*}
Reg(T;\mathbf{u},q)\leq C_{1}N^{2}\sqrt{NT\log TR}+C_{2}N\sqrt{T\log TR\cdot \log T}+\frac{C_{3}N}{R}.
\end{eqnarray*}
\section{Comparison with UCB-V algorithm}
\label{compucbv}
In this section we compare SBORS with UCB-V~\cite{audibert2009exploration} due to the similarities in the way both these techniques maintain estimated means and variances ($\hat{u}_i(t),\hat{q}(t),\hat{\sigma}_{u_i}(t)$ and $\hat{\sigma}_{q}(t)$). The UCB-V algorithm, designed for the vanilla MAB setting, takes the variance of the different arms into consideration while choosing the next action. By estimating the variance explicitly, UCB-V has the ability to reduce the exploration (bonus) budget spent on certain arms, drastically reducing the regret incurred. In particular, it can be shown that the regret of UCB-V is smaller if the variance of suboptimal items is small. 

Although UCB-V algorithm shares some similarities with SBORS algorithm since both these consider variance of the parameters involved, they are fundamentally different. In the SBORS algorithm, parameters $\mathbf{u}, q$ are random variables that are sampled from Gaussian distributions, whereas for the UCB-V algorithm, these are fixed unknowns and their estimates are maintained as $\hat{u}_{i}, \hat{q}$. SBORS achieves exploration via sampling, whereas UCB-V achives exploration via explicit bonus terms and does not rely on randomization.

Nonetheless, we design an extension of UCB-V that uses variance estimates to improve recommendations in our setting based on ideas from \cite{aroyo2018reports} and \cite{audibert2009exploration}. This algorithm (Algorithm \ref{UCBV}) is different from the algorithm proposed by \cite{aroyo2018reports} in that it considers the variance of the parameters related to different items, as shown in Equation (\ref{ucbvu}). The update for $q$ (\ref{ucbvq}) is left unchanged:

\begin{eqnarray}
\label{ucbvu}
u_{i,t}^{UCB}=\hat{u}_{i}(t)+\sqrt{\frac{2\textrm{Var}(\hat{u}_{i}(t))\log t}{T_{i}(t)}}+\frac{b\log t}{T_{i}(t)},
\end{eqnarray}
and 
\begin{eqnarray}
\label{ucbvq}
q_{t}^{UCB}=\hat{q}(t)+\sqrt{\frac{2\log t}{N_{q}(t)}},
\end{eqnarray}
where $\hat{u}_{i}(t)$, $\hat{q}(t)$ can be computed by Lemma \ref{unbiasedes}, $\textrm{Var}(\hat{u}_{i}(t))$ is the estimated variance of $\hat{u}_{i}(t)$ at time $t$, and $b$ is the upper bound on the support of $u_i$s.

\begin{algorithm}[ht!]
\caption{UCB-V algorithm}\label{UCBV}
\begin{algorithmic}
    \STATE {\bfseries Initialization:}
    Set $u_{i,0}^{UCB}=1$ for all $i\in [N]$ and $q_{0}^{UCB}=1$. Set $c_{i}(t)=f_{i}(t)=1$ for all $i\in [N]$, $n_{e}(t)=n_{a}(t)=1$; and $t=1$.

        \WHILE{$t\leq T$}
        
        \STATE{Compute $\mathbf{S}^{t}=\underset{\mathbf{S}}{\arg\max} \expected[U(\mathbf{S};\mathbf{u}_{t-1}^{UCB},q_{t-1}^{UCB})]$ according to Theorem \ref{theorder}.}

        \STATE{Offer sequence $\mathbf{S}^{t}$, observe feedback of user who sees $k_{t} \leq |\mathbf{S}^{t}|$ items.}

        \FOR {$i=1,\cdots,[N]$}
        
        \item[]{Update $u_{I(i),t}^{UCB}$ according to Equation (\ref{ucbvu}).}
        
        \ENDFOR
        
        \item[]{Update $c_{i}(t)$, $f_{i}(t)$, $ne(t)$ and $na(t)$.}
        
        \item[]{Update $q^{UCB}$ according to Equation (\ref{ucbvq}).}
        
        $t=t+1$.
        \ENDWHILE
\end{algorithmic}
\end{algorithm}

\section{Experiments}
\label{experiments}
In this section, we demonstrate the robustness of Algorithm \ref{preSBORS} and Algorithm \ref{sbors} by comparing how the regret changes with respect to different values of $\mathbf{u}$ and other relevant parameters. We also compare our algorithms the UCB-based algorithm of \cite{aroyo2018reports}.

\subsection{Robustness of precursor to SBORS (Algorithm \ref{preSBORS})}
\label{robustnessofalgorithm}

\textbf{Setting:} $N=30$, reward $r_{i}$ is uniformly distributed between $[0, 1]$, abandonment distribution probability $p=0.1$ and the cost of abandonment $c=0.5$. We present four scenarios, when the preference parameter $\mathbf{u}$ is uniformly generated from $[0, 0.1]$, $[0, 0.2]$, $[0, 0.3]$, $[0, 0.5]$, element-wise.

\begin{figure}[ht!]
    \centering
    \subfigure[]{
    \includegraphics[width=5.7cm]{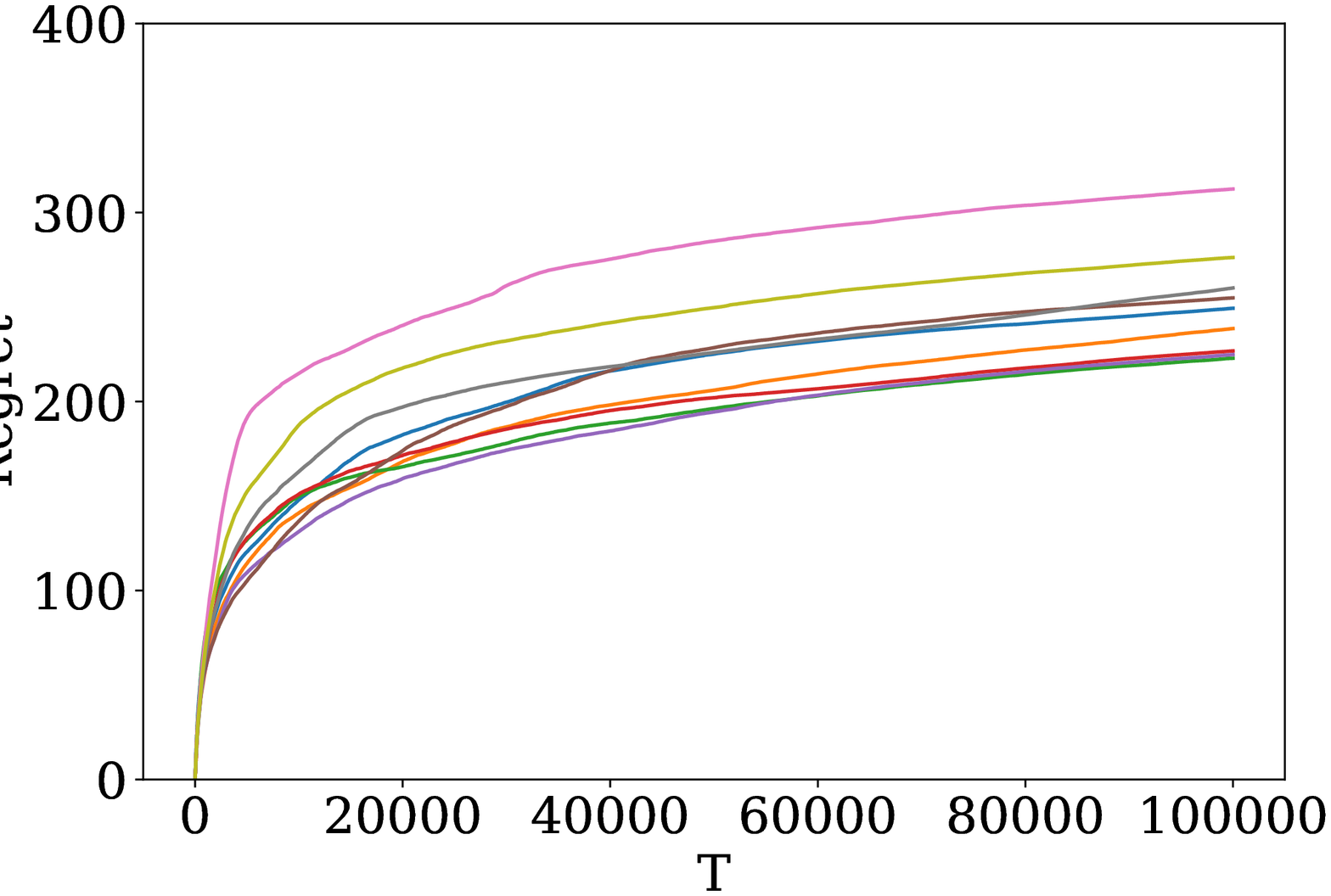}
    }
    \quad
    \subfigure[]{
    \includegraphics[width=5.7cm]{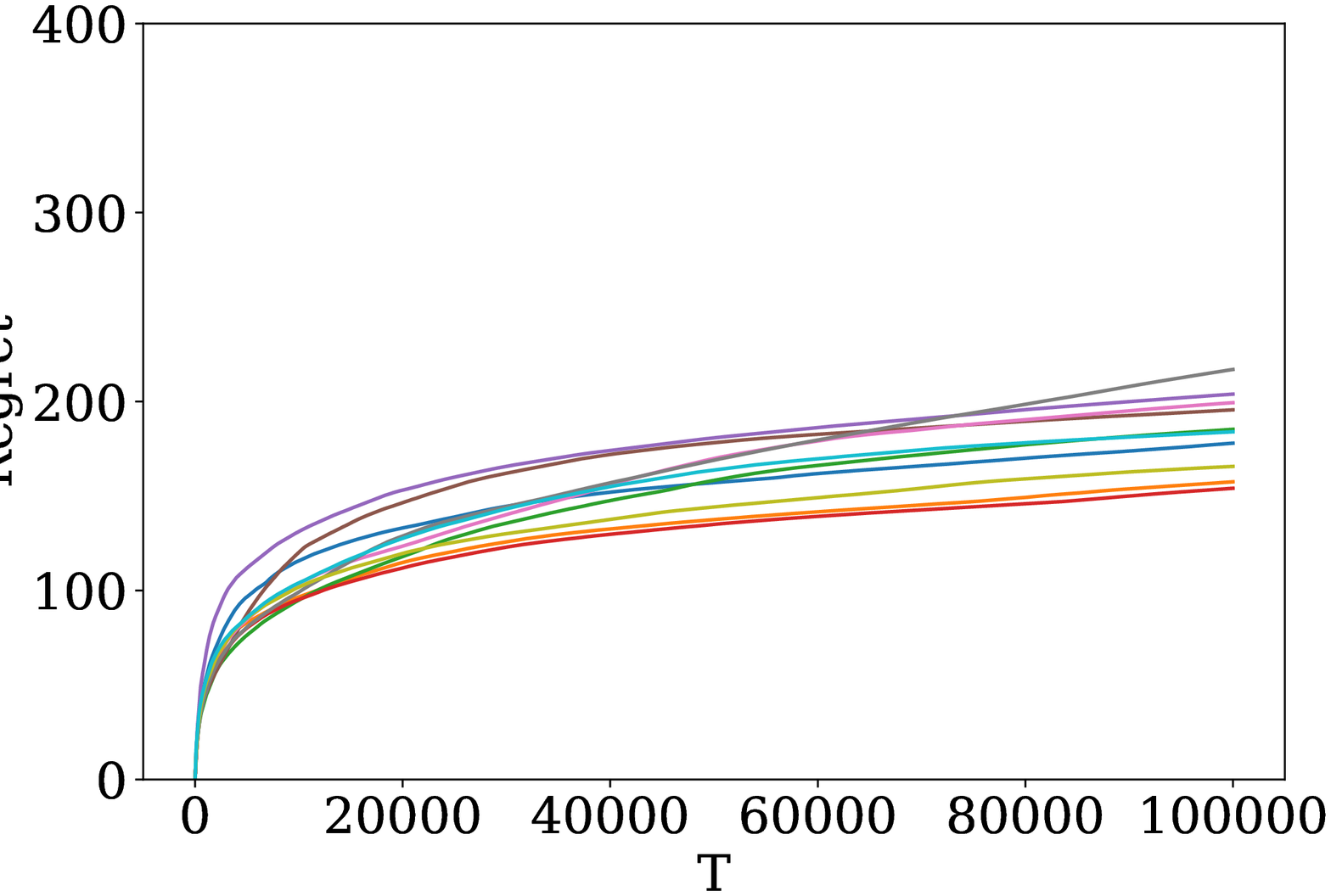}
    }
    \quad
    \subfigure[]{
    \includegraphics[width=5.7cm]{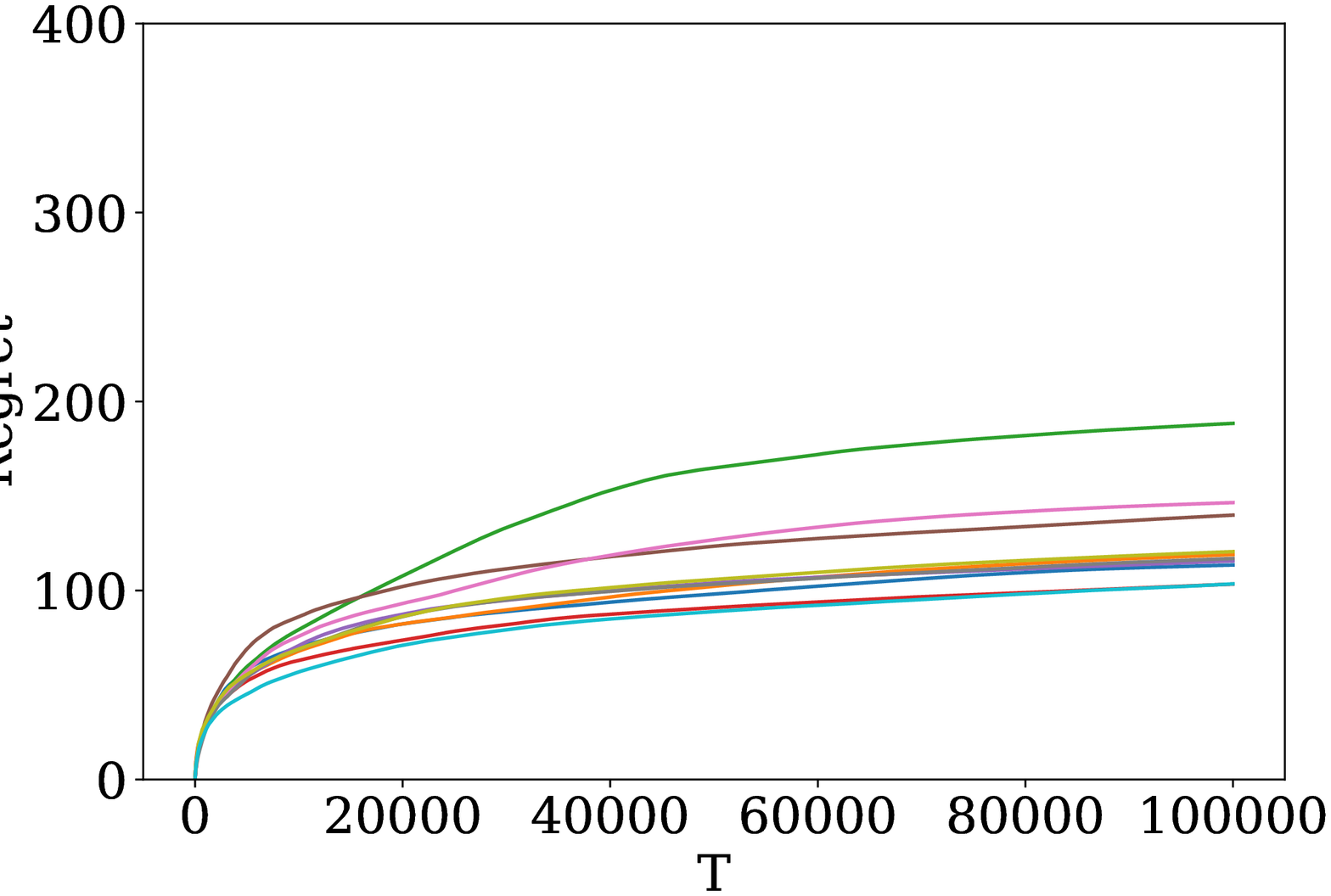}
    }
    \quad
    \subfigure[]{
    \includegraphics[width=5.7cm]{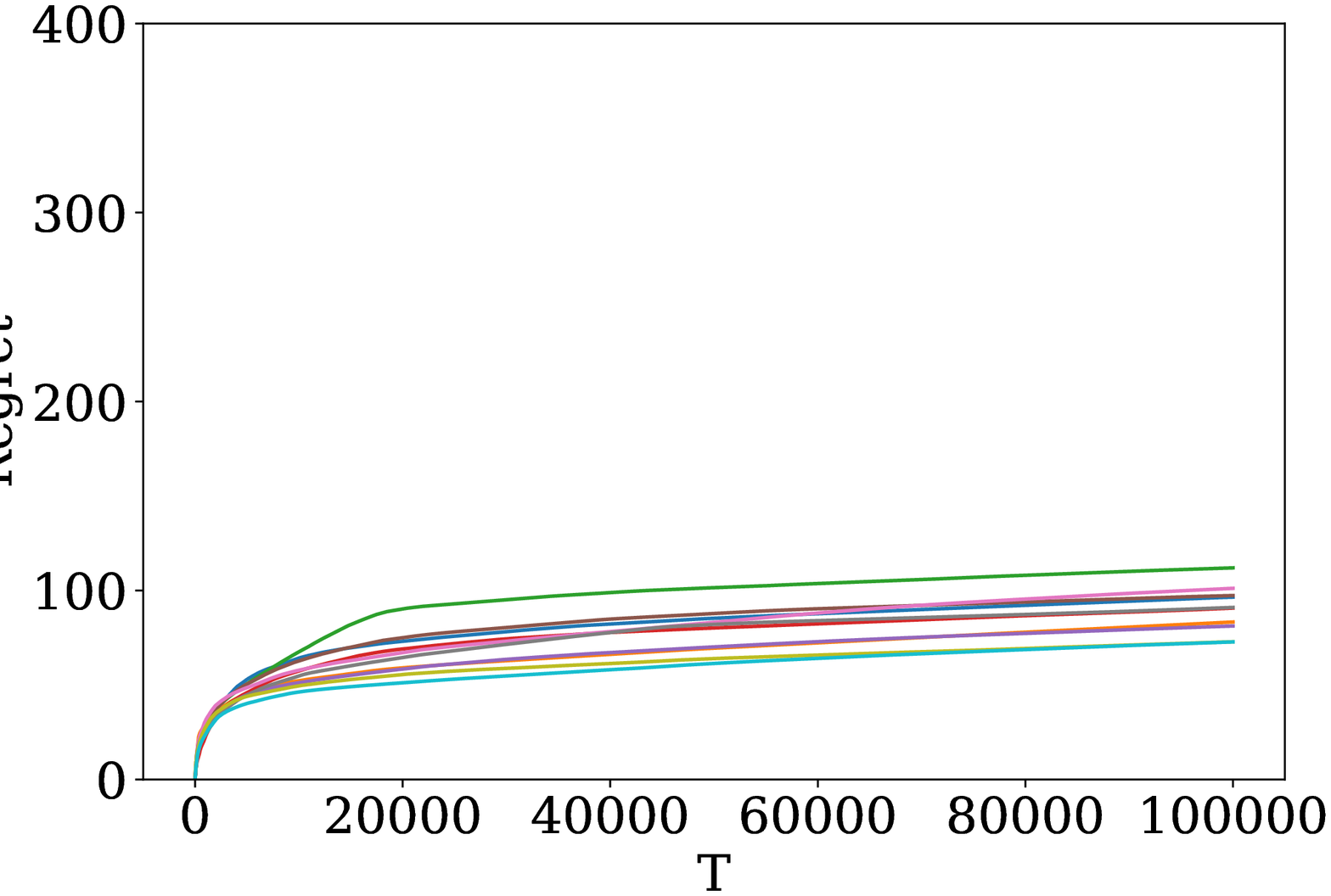}
    }
    \caption{Comparison of Algorithm \ref{preSBORS} when $\mathbf{u}$ is uniformly generated from (a) [0,0.1], (b) [0,0.2], (c) [0,0.3], and (d) [0,0.5].}
    \label{figunit}
\end{figure}

\noindent \textbf{Results:} Figure \ref{figunit} shows the results based on 10 independent simulations for different scenarios of $\mathbf{u}$. The average regrets are $270.1, 186.1, 126.2, 91.2$, respectively. According to figure \ref{figunit}, the regrets eventually tend to stop growing steeply. When $\mathbf{u}$ is generated from $[0, 0.1]$ and $[0, 0.2]$, regret continues to increase after the initial $100,000$ iterations. On the other hand, when $\mathbf{u}$ is generated from $[0, 0.3]$ and $[0, 0.5]$, it converge quickly, for instance before 50,000 and 25,000 rounds respectively. Thus we conclude that the more spread out $\mathbf{u}$ is, the shorter time the algorithm needs to find the optimal sequence, and regret is lower. 

\subsection{Robustness of SBORS (Algorithm \ref{sbors})}

The setting is the same as Section \ref{robustnessofalgorithm}. Additionally, we generate $\mathbf{u}$ form [0,0.1], and discuss the influence of sampling parameter $R$, and fixed constants $\alpha, \beta$ on the regret separately.

\begin{figure}[ht!]
    \centering
    \subfigure[$u_i$ is uniformed generated from  $0$ to $0.1, 0.2, 0.3, 0.5$, respectively.]{
    \includegraphics[width=5.7cm]{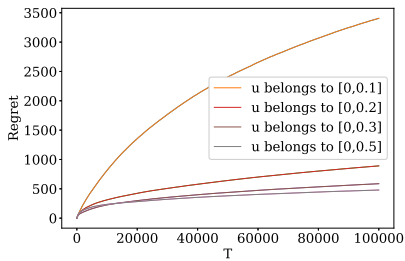}
    \label{figcomptsgu}
    }
    \quad
    \subfigure[$R$ is 1, 10, 100, respectively.]{
    \includegraphics[width=5.7cm]{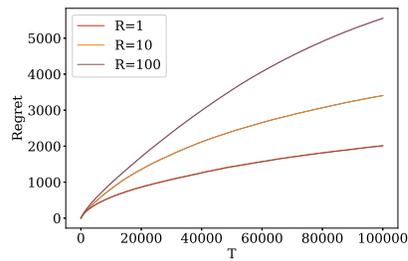}
    \label{figcomptsgR}
    }
    \quad
    \subfigure[$\alpha$ is 0.1, 1, 10, respectively.]{
    \includegraphics[width=5.7cm]{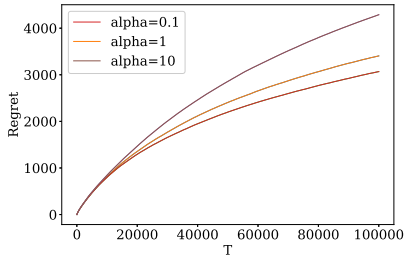}
    \label{figcomptsgalpha}
    }
    \quad
    \subfigure[$\beta$ is 0.2, 2, 20, respectively.]{
    \includegraphics[width=5.7cm]{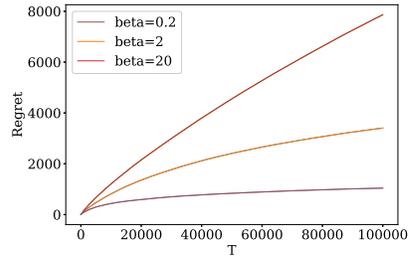}
    \label{figcomptsgbeta}
    }
    \caption{Plots for SBORS with different $\mathbf{u}$, $R$, $\alpha$, and $\beta$.}
    \label{figts}
\end{figure}

\textbf{Influence of $\mathbf{u}$} (Figure \ref{figcomptsgu}): We can infer that the more $\mathbf{u}$ is spread out, the lower the regret is, which is in agreement with Figure \ref{figunit}.

\textbf{Influence of $R$} (Figure \ref{figcomptsgR}): We set $\alpha=1$, $\beta=2$ and vary $R$. We can infer that lower $R$ values reduce the regret. One extreme case is $R=1$, which essentially removes variance boosting and still performs well empirically.

\textbf{Influence of $\alpha$} (Figure \ref{figcomptsgalpha}): We set $R=10$, $\beta=2$ and change $\alpha$. We can infer that lower $\alpha$ values reduce the regret. 

\textbf{Influence of $\beta$} (Figure \ref{figcomptsgbeta}): We set $R=10$, $\alpha=1$ and change $\beta$. We can infer that lower $\beta$s reduce regret. For analysis, we needed $\beta \geq 2$, but we observe that choosing $\beta<2$ can still lead to better regret hinting at a potential slack in our analysis.

\subsection{Comparison with benchmark algorithms}
\begin{figure}[ht!]
    \centering
    \subfigure[UCB-based algorithm]{
    \includegraphics[width=5.7cm]{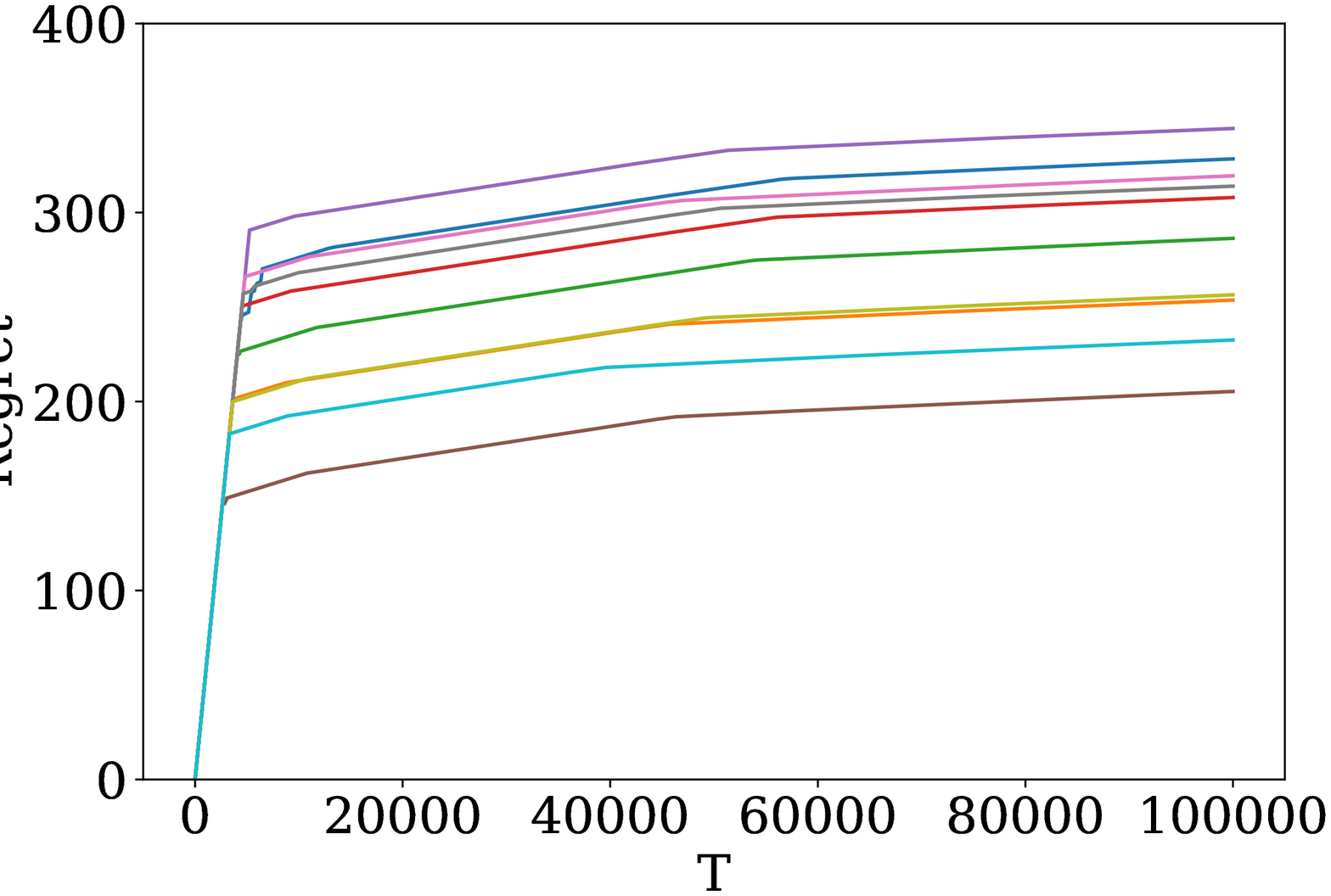}
    }
    \quad
    \subfigure[UCB-V algorithm]{
    \includegraphics[width=5.7cm]{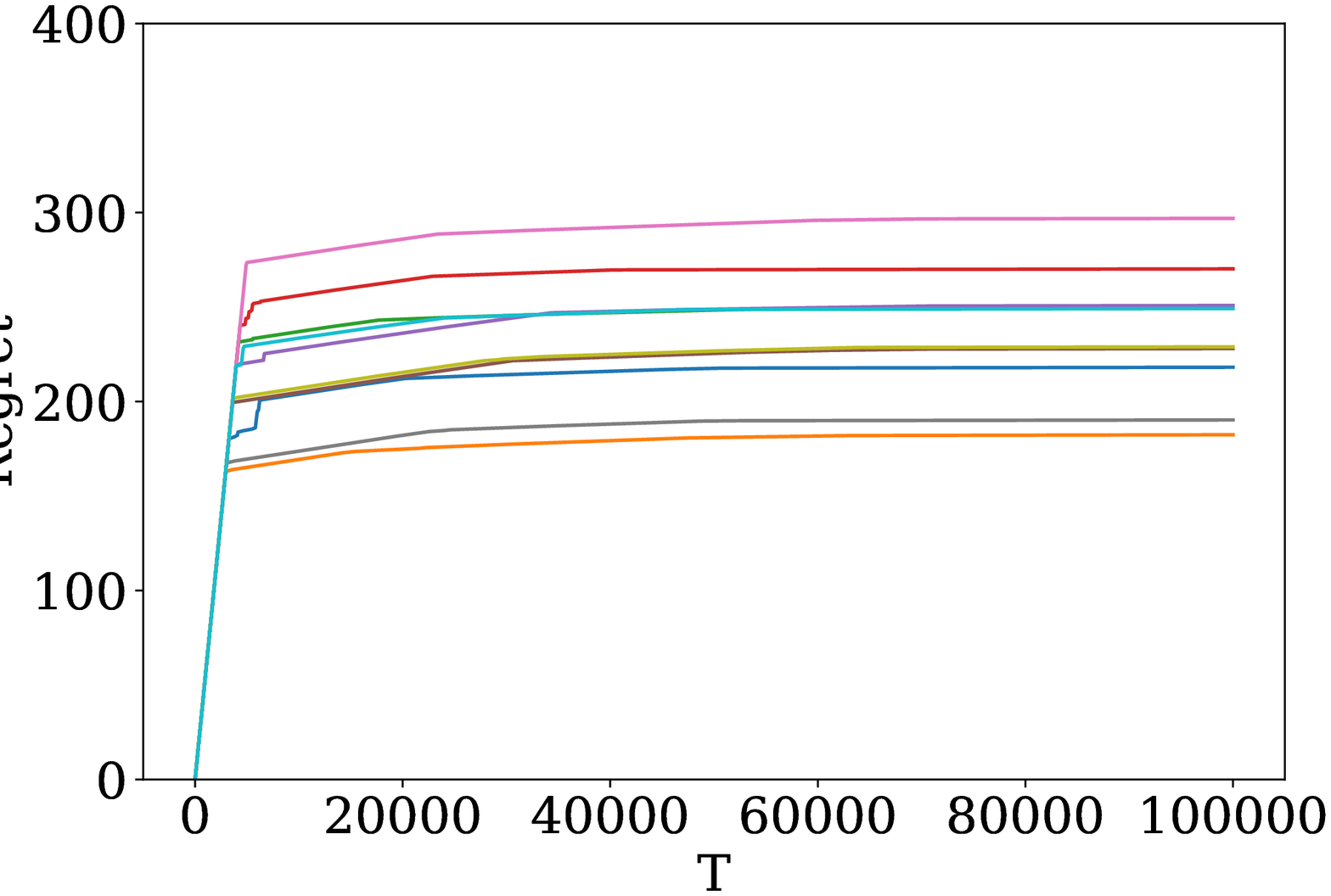}
    }
    \quad
    \subfigure[Our algorithm]{
    \includegraphics[width=5.7cm]{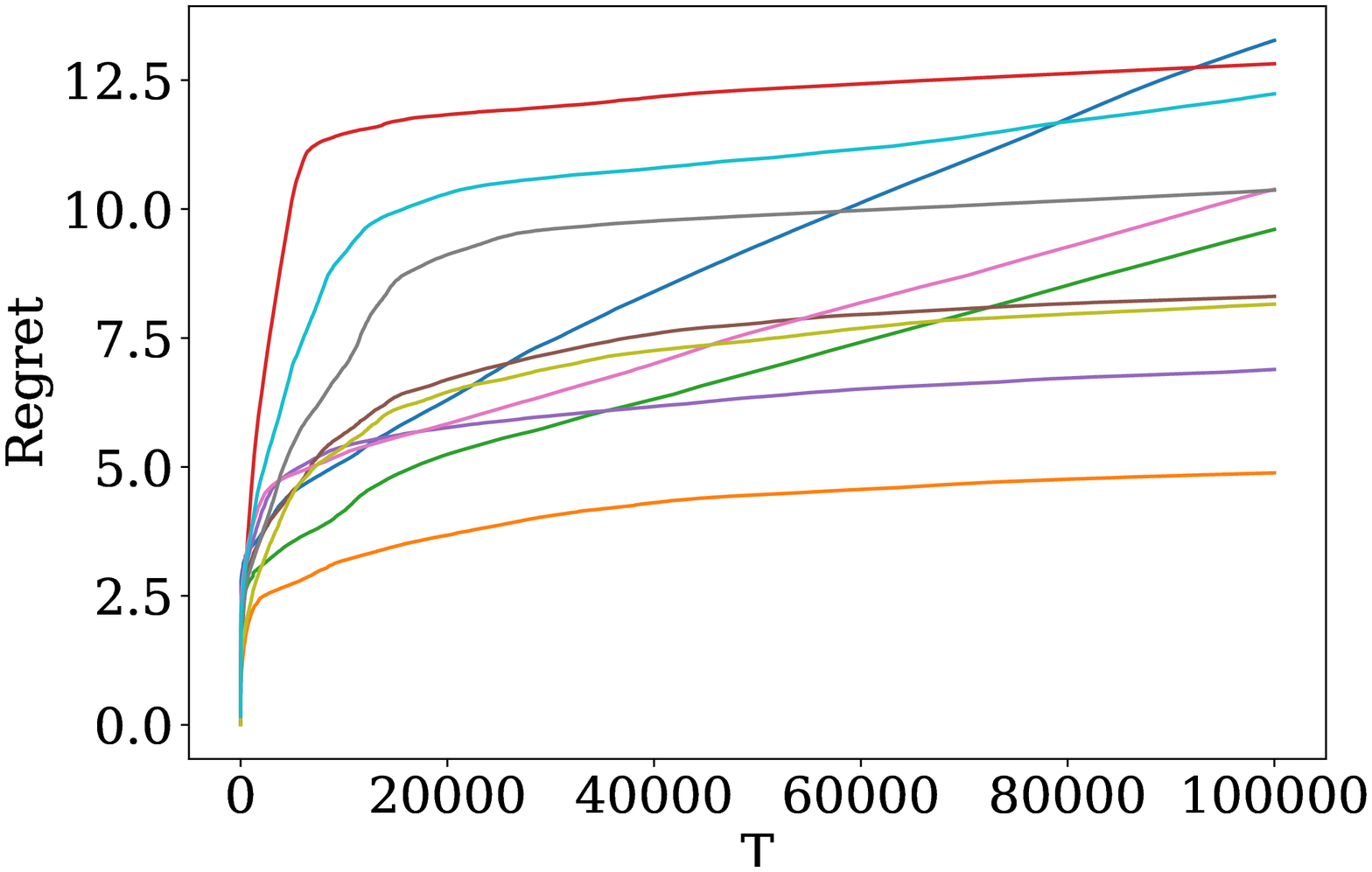}
    }
    \caption{Comparison of UCB-based algorithm, UCB-V algorithm and Algorithm~\ref{preSBORS}.}
    \label{figcompucbts}
\end{figure}

We compare Algorithm \ref{preSBORS} with the algorithm in \cite{aroyo2018reports} and its UCB-V variant (Algorithm \ref{UCBV}) defined in Section \ref{compucbv}. The setting is the same as in Section \ref{robustnessofalgorithm}, except we only present results for $\mathbf{u}$ uniformly generated from $[0,1]$. Figure \ref{figcompucbts} shows the cumulative regrets incurred using the three algorithms separately over multiple runs. It suggests that the regret of our algorithm is much lower (a factor of $5\times$ or more) compared to the UCB-based and the UCB-V algorithms. 

\section{Conclusion}
\label{conclusion}
In this paper, we present a new Thompson sampling based algorithm for making recommendations where users experience fatigue. We use techniques such as posterior approximation using Gaussians, correlate sampling and variance boosting to control the exploration-exploitation trade-off and derive rigorous regret upper bounds. Our bounds depend polynomially on the number of items and sub-linearly on the time horizon ($C_{1}N^{2}\sqrt{NT\log TR}+C_{2}N\sqrt{T\log TR\cdot \log T}+C_{3}N/R$). Our algorithm outperforms UCB-based approaches in simulations and can be easily extended to contextual settings. Future directions include tackling the computational complexity of the combinatorial problem in each round, tightening the regret upper bound, and extending the machinery to recommendation systems with a variety of other user behavior models.

\bibliographystyle{informs2014} % outcomment this and next line in Case 1
\bibliography{reference} % if more than one, comma separated

\section*{Appendix A: Proofs of Lemmas}

\noindent\textbf{Proof of Lemma \ref{unbiasedes}:} The detail can be seen in \cite{aroyo2018reports} Proof of Lemma 5.

\noindent\textbf{Proof of Lemma \ref{lm}:}

We first proof equation 1 by using Hoeffding's inequality, which is shown below:
\begin{eqnarray*}
P\left(|\hat{u}_{i}(t)-u_{i}|\geq
\sqrt{\frac{\beta\log \rho}{T_{i}(t)}}\right)
\leq 2e^{-2\beta\log\rho}
=\frac{2}{\rho^{2\beta}}
\end{eqnarray*}

Since $\sqrt{\frac{\alpha\hat{u}_{i}(t)(1-\hat{u}_{i}(t))\log\rho}{T_{i}(t)+1}}$ 
%and $\sqrt{\frac{\alpha\hat{q}(t)(1-\hat{q}(t))\log \rho}{N_{q}(t)+1}}$ are both 
is greater than 0, we have that
\begin{eqnarray*}
\begin{aligned}
&P\left(|\hat{u}_{i}(t)-u_{i}|\geq\sqrt{\frac{\alpha\hat{u}_{i}(t)(1-\hat{u}_{i}(t))\log\rho}{T_{i}(t)+1}}+
\sqrt{\frac{\beta\log \rho}{T_{i}(t)}}\right)\\
&\leq P\left(|\hat{u}_{i}(t)-u_{i}|\geq
\sqrt{\frac{\beta\log \rho}{T_{i}(t)}}\right)\\
&\leq\frac{2}{\rho^{2\beta}}
\end{aligned}
\end{eqnarray*}

Similarly, we can prove equation 2 by replacing $\hat{u}_{i}(t)$ with $\hat{q}(t)$, $u_{i}$ with $q$, $T_{i}(t)$ with $N_{q}(t)$.

\noindent\textbf{Proof of Lemma \ref{anticon}:}

Note that we have $u_{i}'(t)=\hat{u}_{i}(t)+\hat{\sigma}_{u_{i}}(t)\cdot \underset{j=1,\cdots,R}{max}\{\theta^{(j)}(t)\}$, $q'(t)=\hat{q}(t)+\hat{\sigma}_{q}(t)\cdot \underset{j=1,\cdots,R}{max}\{\theta^{(j)}(t)\}$. Therefore, from union bound, we have,
\begin{eqnarray*}
\begin{aligned}
P\left\{|u_{i}'(t)-\hat{u}_{i}(t)|>4\hat{\sigma}_{u_{i}}(t)\sqrt{\log rR}|\hat{u}_{i}(t)\right\}
&=P\left(\bigcup_{j=1}^{R}\{\theta^{(j)}(t)>4\sqrt{\log rR}\}\right)\\
&\leq \sum_{j=1}^{R}P\left(\theta^{(j)}(t)>4\sqrt{\log rR}\right)\\
&\overset{a_{1}}{=}\frac{1}{r^{7}R^{6}}\cdot \frac{1}{4\sqrt{\log rR}}\\
&\overset{a_{2}}{\leq} \frac{1}{r^{7}R^{6}}
\end{aligned}
\end{eqnarray*}

Similarly,
\begin{eqnarray*}
P\left\{|q'(t)-\hat{q}(t)|>4\hat{\sigma}_{q}(t)\sqrt{\log rR}|\hat{q}(t)\right\}\leq \sum_{j=1}^{R}P\left(\theta^{(j)}(t)>4\sqrt{\log rR}\right)
\leq \frac{1}{r^{7}R^{6}}
\end{eqnarray*}

Equality $(a_{1})$ can be calculated from the tail bound for Gaussian random variables $\theta^{(j)}(t)$.
\begin{eqnarray*}
%\frac{1}{4\sqrt{\pi}}\cdot e^{-\frac{7z^{2}}{2}}<
P(|\theta^{(j)}(t)|>z)\leq \frac{1}{2z}e^{-\frac{z^{2}}{2}}
\end{eqnarray*}

Inequality $(a_{2})$ holds because $r>1$ and $R>1$, then $\frac{1}{4\sqrt{\log rR}}<rR$.

\noindent\textbf{Proof of Lemma \ref{lip}:}

\emph{Proof 1:}
Please see the proof of Lemma 7 in \cite{aroyo2018reports}.

\emph{Proof 2:}

For any message sequence $\mathbf{S}$ of length $m$, let $\mathbf{S}_{j}$ be the sub-sequence starting from the $j^{th}$ message, i.e. $\mathbf{S}_{j}=(S_{j},S_{j+1},\cdots,S_{m})$. 

Define event $C_{j}$ as a user views the $j^{th}$ message in the sequence, $\expected[U(\mathbf{S}_{j};\mathbf{u},q_u)|C_{j}]$ as the partial expected payoff conditioned on a user viewing the $j^{th}$ message in the sequence. For explicitness, we use $PE[U(\mathbf{S}_{j};\mathbf{u},q_{u})]$ instead. From Section 3, we can recall that $P(W=i|W\geq j)$ means the probability that the user abandons the platform after s/he sees message $i$ on the condition that the number of unsatisfied messages the user has seen is no less than $j$. Let $I(\cdot)$ be an identity function for simplicity, we have
\begin{eqnarray*}
\begin{aligned}
PE[U(\mathbf{S}_{j};\mathbf{u},q_{u})]&=r_{j}u_{j}-c(1-u_{j})P(W=j|W\geq j)\\
&+\sum_{i=j+1}^{|\mathbf{S}|}\bigg(\prod_{k=j}^{i-1}(1-u_{k})\bigg)\bigg(r_{i}u_{i}P(W\geq i|W\geq j)-c(1-u_{i})P(W=i|W\geq j)\bigg)\\
&=r_{j}u_{j}-pc(1-u_{j})+\sum_{i=j+1}^{|\mathbf{S}|}q^{i-j}\bigg(\prod_{k=j}^{i-1}(1-u_{k})\bigg)\big(r_{i}u_{i}-pc(1-u_{i})\big)\\
&=r_{j}u_{j}+(1-u_{j})(qPE[U(\mathbf{S}_{j+1};\mathbf{u},q_{u})]-pc)
\end{aligned}
\end{eqnarray*}

Define vector $\mathbf{u}$ and $q_{u}$ as
\begin{eqnarray*}
u_{i}=\max\{v_{i},w_{i}\}\\
q_{u}=\max\{q_{v},q_{w}\}
\end{eqnarray*}

Denote $\mathbf{S}_{j}^{v*}$ as the sub-sequence of the optimal sequence message $\mathbf{S}_{v}^{*}$ with parameter $\mathbf{v}$ and $q_{v}$ starting from the $j^{th}$ message.

Therefore, %from the restricted monotonicity property, 
we have
\begin{eqnarray}
\begin{aligned}
\label{pein}
%& \expected[U(\mathbf{S}^{*}_{j},\mathbf{v},q_{v})]-\expected[U(\mathbf{S}^{*}_{j},\mathbf{w},q_{w})]\\
&PE[U(\mathbf{S}^{v*}_{j};\mathbf{u},q_{u})]-PE[U(\mathbf{S}^{v*}_{j};\mathbf{w},q_{w})]\\
&=r_{j}u_{j}+(1-u_{j})(q_{u}PE[U(\mathbf{S}_{j+1}^{v*};\mathbf{u},q_{u})]-(1-q_{u})c)\\
&\quad\quad-r_{j}w_{j}-(1-w_{j})(q_{w}PE[U(\mathbf{S}_{j+1}^{v*};\mathbf{w},q_{w})]-(1-q_{w})c)\\
&= r_{j}(u_{j}-w_{j})-c\big((1-q_{u})(1-u_{j})-(1-q_{w})(1-w_{j})\big)\\
&\quad\quad+\big(q_{u}(1-u_{j})PE[U(\mathbf{S}_{j+1}^{v*};\mathbf{u},q_{u})]-q_{w}(1-w_{j})PE[U(\mathbf{S}_{j+1}^{v*};\mathbf{w},q_{w})]\big)\\
&\overset{b_{1}}{\leq} r_{j}(u_{j}-w_{j})+c(q_{u}-q_{w}+u_{j}-w_{j})\\
&\quad\quad+\big(q_{u}(1-u_{j})PE[U(\mathbf{S}_{j+1}^{v*};\mathbf{u},q_{u})]-q_{w}(1-w_{j})PE[U(\mathbf{S}_{j+1}^{v*};\mathbf{w},q_{w})]\big)\\
&\overset{b_{2}}{\leq} (r_{j}+c)(u_{j}-w_{j})+c(q_{u}-q_{w})\\
&\quad\quad+\big(q_{u}(1-u_{j})PE[U(\mathbf{S}_{j+1}^{v*};\mathbf{u},q_{u})]-q_{w}(1-u_{j})PE[U(\mathbf{S}_{j+1}^{v*};\mathbf{w},q_{w})]\big)\\
&\overset{b_{3}}{\leq} (r_{j}+c)(u_{j}-w_{j})+c(q_{u}-q_{w})
+\big|q_{u}PE[U(\mathbf{S}_{j+1}^{v*};\mathbf{u},q_{u})]-q_{w}PE[U(PS_{j+1}^{v*};\mathbf{w},q_{w})]\big|\\
&\overset{b_{4}}{\leq} 2(u_{j}-w_{j})+(q_{u}-q_{w})
+\big|q_{w}(PE[U(\mathbf{S}_{j+1}^{v*};\mathbf{u},q_{u})]-PE[U(PS_{j+1}^{v*};\mathbf{w},q_{w})])\\
&\quad\quad+(q_{u}-q_{w})PE[U(\mathbf{S}_{j+1}^{v*};\mathbf{u},q_{u})]\big|\\
&\overset{b_{5}}{\leq} 2(u_{j}-w_{j})+(q_{u}-q_{w})
+\big|PE[U(\mathbf{S}_{j+1}^{v*};\mathbf{u},q_{u})]-PE[U(PS_{j+1}^{v*};\mathbf{w},q_{w})]\big|\\
&\quad\quad+(q_{u}-q_{w})PE[U(\mathbf{S}_{j+1}^{v*};\mathbf{u},q_{u})]\\
&\overset{b_{6}}{\leq}2(u_{j}-w_{j})+(N+1)(q_{u}-q_{w})
+\big|PE[U(\mathbf{S}_{j+1}^{v*};\mathbf{u},q_{u})]-PE[U(PS_{j+1}^{v*};\mathbf{w},q_{w})]\big|
\end{aligned}
\end{eqnarray}

Inequality $(b_{1})$ follows from the observation that
\begin{eqnarray*}
\begin{aligned}
&(1-q_{u})(1-u_{j})-(1-q_{w})(1-w_{j})\\
&=(1-q_{u}-u_{j}+q_{u}u_{j})-(1-q_{w}-w_{j}+q_{w}w_{j})\\
&=q_{u}u_{j}-q_{w}w_{j}-(q_{u}-q_{w})-(u_{j}-w_{j})\\
&\geq -(q_{u}-q_{w})-(u_{j}-w_{j})
\end{aligned}
\end{eqnarray*}

Inequality $(b_{2})$ is because we replace $w_{j}$ with $u_{j}$.

Inequality $(b_{3})$ holds because $0\leq 1-u_{j}\leq 1$.

Inequality $(b_{4})$ is because we add and subtract $q_{w}\expected[U(\mathbf{S}_{j+1}^{*};\mathbf{u},q_{u})]$ and $r_{j}, c\in [0,1]$.

Inequality $(b_{5})$ holds because $q_{w}\in[0,1]$, $q_{u}\geq q_{w}$, $PE[U(\mathbf{S}_{j+1}^{v*};\mathbf{w},q_{w})]\geq 0$ and absolute value property $|a+b|\leq |a|+|b|$.

Inequality $(b_{6})$ holds because of definition that 
\begin{eqnarray*}
\begin{aligned}
\mathbf{S}_{1}^{v*}&=\mathbf{S}^{*}_{v}\\
PE[U(\mathbf{S}_{1}^{v*};\mathbf{u},q_{u})]
&=\expected[U(\mathbf{S}^{*}_{v};\mathbf{u},q_{u})]\\
PE[U(\mathbf{S}_{j+1}^{v*};\mathbf{u},q_{u})]
&\leq PE[U(\mathbf{S}_{1}^{v*};\mathbf{u},q_{u})]\\
&=\expected[U(\mathbf{S}^{*}_{v};\mathbf{u},q_{u})]\\
&=\sum_{i\in \mathbf{S}^{*}_{v}}p_{i}(\mathbf{S}^{*}_{v})r_{i}-cp_{a}(\mathbf{S}^{*}_{v})\textrm{  (by definition)}\\
&\leq N
\end{aligned}
\end{eqnarray*}

Since
\begin{eqnarray*}
PE[U(\mathbf{S}_{m},\mathbf{u},q_{u})]=r_{m}u_{m}-c(1-q_{u})(1-u_{m})
\end{eqnarray*}

We have,
\begin{eqnarray*}
\begin{aligned}
&PE[U(\mathbf{S}_{m}^{v*};\mathbf{u},q_{u})]-PE[U(\mathbf{S}_{m}^{v*};\mathbf{w},q_{w})]\\
&=r_{m}(u_{m}-w_{m})-c\left((1-q_{u})(1-u_{m})-(1-q_{w})(1-w_{m})\right)\\
&\leq (r_{m}+c)(u_{m}-w_{m})+c(q_{u}-q_{w})\\
&\leq 2(u_{m}-w_{m})+(q_{u}-q_{w})\\
&\leq 2(u_{m}-w_{m})+(N+1)(q_{u}-q_{w})
\end{aligned}
\end{eqnarray*}

According to inequality (\ref{pein}), by induction, we can get
\begin{eqnarray*}
\begin{aligned}
\left|\expected[U(\mathbf{S}^{*}_{v};\mathbf{v},q_{v})]-\expected[U(\mathbf{S}^{*}_{v};\mathbf{w},q_{w})]\right|&\overset{b_{7}}{\leq} \expected[U(\mathbf{S}^{*}_{v};\mathbf{u},q_{u})]-\expected[U(\mathbf{S}^{*}_{v};\mathbf{w},q_{w})]\\
&=PE[U(\mathbf{S}^{v*}_{1};\mathbf{u},q_{u})]-PE[U(\mathbf{S}^{v*}_{1};\mathbf{w},q_{w})]\\
&\leq \sum_{i \in \mathbf{S}^{*}_{v}}\left(2(u_{j}-w_{j})+(N+1)(q_{u}-q_{w})\right)\\
&\leq\sum_{i \in \mathbf{S}^{*}_{v}}\left(2|v_{j}-w_{j}|+(N+1)|q_{v}-q_{w}|\right)
\end{aligned}
\end{eqnarray*}

Inequality $(b_{7})$ holds because of the restricted monotonicity in Lemma \ref{lip} part 1.

\noindent\textbf{Proof of Lemma \ref{disexp}:}

\noindent Notations:
\begin{itemize}
    \item For any $t\leq T$, define $\Delta U_{t}$ as follows,
    \begin{eqnarray*}
    \Delta U_{t}\overset{\Delta}{=}\expected[U(\mathbf{S}^{t};\mathbf{u}'(t),q'(t))]-\expected[U(\mathbf{S}^{t};\mathbf{u},q)]
    \end{eqnarray*}
    
    \item For any $t\in \{1,\cdots,T\}$, define events $\mathcal{G}_{t}$, $\mathcal{H}_{t}$ as
    \begin{eqnarray*}
    \mathcal{G}_{t}=\left\{|\hat{u}_{i}(t)-u_{i}|\geq \sqrt{\frac{\alpha\hat{u}_{i}(t)(1-\hat{u}_{i}(t))\log (t+1)}{T_{i}(t)+1}}+\sqrt{\frac{\beta\log (t+1)}{T_{i}(t)}}\textrm{ for some } i=1,\cdots,N\right\}
    \end{eqnarray*}

    \begin{eqnarray*}
    \mathcal{H}_{t}=\left\{|\hat{q}(t)-q|\geq \sqrt{\frac{\alpha\hat{q}(t)(1-\hat{q}(t))\log (t+1)}{N_{q}(t)+1}}+\sqrt{\frac{\beta\log (t+1)}{N_{q}(t)}}\right\}
    \end{eqnarray*}
    where the definition of $\hat{\sigma}_{u_{i}}(t)$ and $\hat{\sigma}_{q}(t)$ can be seen in Algorithm \ref{sbors}.

    \item Define events $\mathcal{A}_{t}=\mathcal{G}_{t}\cap \mathcal{H}_{t}$
\end{itemize}

Since $\mathcal{A}_{t}$ is a ``low probability" event, we analyze the expected payoff of $\Delta U_{t}$ in two senarios, one when $\mathcal{A}_{t}$ is true and another when $\mathcal{A}_{t}^{c}$ is true. More specifically,
\begin{eqnarray*}
\expected[\Delta U_{t}]=\expected[\Delta U_{t}\cdot \mathbbm{1}(\mathcal{A}_{t})+\Delta U_{t}\cdot \mathbbm{1}(\mathcal{A}_{t}^{c})]
\end{eqnarray*}

Substituting $\rho=t+1$ in Lemma \ref{lm}, we obtain that $P(\mathcal{A}_{t})\leq \frac{2N}{(t+1)^{2\beta}}\times \frac{2}{(t+1)^{2\beta}}\leq\frac{4N}{t^{4\beta}}$. Therefore, it follows that,
\begin{eqnarray*}
\expected[\Delta U_{t}]\leq \frac{4N}{t^{4\beta}}+\expected[\Delta U_{t}\cdot \mathbbm{1}(\mathcal{A}_{t}^{c})]
\end{eqnarray*}

Consider function $I(\cdot)$ is a identity function. From Lemma \ref{lip}, we have that
\begin{eqnarray*}
\begin{aligned}
\Delta U_{t}\leq \sum_{i \in \mathbf{S}^{t}}\left(2|u_{i}'(t)-u_{i}|+(N+1)|q'(t)-q|\right)
\end{aligned}
\end{eqnarray*}

Therefore, it follows that,
\begin{eqnarray*}
\begin{aligned}
\expected[\Delta U_{t}\cdot \mathbbm{1}(\mathcal{A}_{t}^{c})]
\leq
\expected\left[\bigg(\sum_{i \in \mathbf{S}^{t}}\left(2|u_{i}'(t)-u_{i}|+(N+1)|q'(t)-q|\right)\bigg)\cdot \mathbbm{1}(\mathcal{A}_{t}^{c})\right]
\end{aligned}
\end{eqnarray*}

From triangle inequality, we have
\begin{eqnarray*}
\begin{aligned}
\expected[\Delta U_{t}\cdot \mathbbm{1}(\mathcal{A}_{t}^{c})]
&\leq \expected\left[\bigg(\sum_{i \in \mathbf{S}^{t}}\big(2|u_{i}'(t)-\hat{u}_{i}(t)|+(N+1)|q'(t)-\hat{q}(t)|\big)\bigg)\cdot \mathbbm{1}(\mathcal{A}_{t}^{c})\right]\\
&+\expected\left[\bigg(\sum_{i \in \mathbf{S}^{t}}\big(2|\hat{u}_{i}(t)-u_{i}|+(N+1)|\hat{q}(t)-q|\big)\bigg)\cdot \mathbbm{1}(\mathcal{A}_{t}^{c})
\right]
\end{aligned}
\end{eqnarray*}

From the definition of the event $\mathcal{A}_{t}^{c}$, it follows that,
\begin{eqnarray}
\label{at1csum}
\begin{aligned}
\expected[\Delta U_{t}\cdot \mathbbm{1}(\mathcal{A}_{t}^{c})]
&\leq 
\expected\left[\bigg(\sum_{i \in \mathbf{S}^{t}}\left(2|u_{i}'(t)-\hat{u}_{i}(t)|+(N+1)|q'(t)-\hat{q}(t)|\right)\bigg)\cdot \mathbbm{1}(\mathcal{A}_{t}^{c})\right]\\
%E\left[\bigg(2\sum_{i \in \mathbf{S}^{t}}|u_{i}'(t)-\hat{u}_{i}(t)|+N^{2}|q'(t)-\hat{q}(t)|\bigg)\cdot \mathbbm{1}(\mathcal{A}_{t}^{c})\right]\\
&+\expected\left[\sum_{i \in \mathbf{S}^{t}}2\big(\sqrt{\frac{\alpha\hat{u}_{i}(t)(1-\hat{u}_{i}(t))\log t}{T_{i}(t)+1}}+\sqrt{\frac{\beta\log t}{T_{i}(t)}}\big)\right]\\
&+\expected\left[\sum_{i \in \mathbf{S}^{t}}(N+1)\big(\sqrt{\frac{\alpha\hat{q}(t)(1-\hat{q}(t))\log t}{N_{q}(t)+1}}+\sqrt{\frac{\beta\log t}{N_{q}(t)}}\big)\right]
\end{aligned}
\end{eqnarray}

We now focus on the bounding the first term in (\ref{at1csum}). In Lemma \ref{anticon}, we show that for any $r>1$, and $i=1,\cdots, N$, we have,
\begin{eqnarray*}
P(|u_{i}'(t)-\hat{u}_{i}(t)|>4\hat{\sigma}_{u_{i}}(t)\sqrt{\log rR})\leq \frac{1}{r^{7}R^{6}}
\end{eqnarray*}
where $\hat{\sigma}_{u_{i}}(t)=\sqrt{\frac{\alpha\hat{u}_{i}(t)(1-\hat{u}_{i}(t))}{T_{i}(t)+1}}+\sqrt{\frac{\beta}{T_{i}(t)}}$

Since $|u_{i}'(t)-\hat{u}_{i}(t)|$ and $|q'(t)-\hat{q}(t)|$ are both non-negative random variables, we have
\begin{eqnarray*}
\begin{aligned}
\expected\left[|u_{i}'(t)-\hat{u}_{i}(t)|\right]&=\int_{0}^{\infty}P\{|u_{i}'(t)-\hat{u}_{i}(t)|\geq x\}dx\\
&=\int_{0}^{4\hat{\sigma}_{u_{i}}(t)\sqrt{\log TR}}P\{|u_{i}'(t)-\hat{u}_{i}(t)|\geq x\}dx+
\int_{4\hat{\sigma}_{u_{i}}(t)\sqrt{\log TR}}^{\infty}P\{|u_{i}'(t)-\hat{u}_{i}(t)|\geq x\}dx\\
&\leq 4\hat{\sigma}_{u_{i}}(t)\sqrt{\log TR}+\sum_{r=T}^{\infty}\int_{4\hat{\sigma}_{u_{i}}(t)\sqrt{\log TR}}^{4\hat{\sigma}_{u_{i}}(t)\sqrt{\log (r+1)R}}P\{|u_{i}'(t)-\hat{u}_{i}(t)|\geq x\}dx\\
&\leq 4\hat{\sigma}_{u_{i}}(t)\sqrt{\log TR}+\sum_{r=T}^{\infty}\frac{4\hat{\sigma}_{u_{i}}(t)}{r^{7}R^{6}}(\sqrt{\log (r+1)R}-\sqrt{\log rR})\\
&\overset{c}{\leq} 4\hat{\sigma}_{u_{i}}(t)\sqrt{\log TR}+4\hat{\sigma}_{u_{i}}(t)\sum_{r=T}^{\infty}\frac{1}{r^{8}R^{6}}\\
%&\leq 4\hat{\sigma}_{u_{i}}(t)\sqrt{\log TR}+4\hat{\sigma}_{u_{i}}(t)\sum_{r=T}^{\infty}\frac{1}{r^{4}R^{3}} \textrm{ for any } T\geq N\\
&\leq 4\hat{\sigma}_{u_{i}}(t)(\sqrt{\log TR}+D_{1})%\textrm{ for any } T\geq N
\end{aligned}
\end{eqnarray*}

Inequality $(c)$ holds because 
\begin{eqnarray*}
\begin{aligned}
\frac{\sqrt{\log (r+1)R}-\sqrt{\log rR}}{r^{7}R^{6}}
&\leq \frac{\log (r+1)R-\log rR}{r^{7}R^{6}(\sqrt{\log (r+1)R}+\sqrt{\log rR})}\\
&\leq \frac{\log (1+\frac{1}{r})}{r^{7}R^{8}}\\
&\leq \frac{1}{r^{8}R^{6}}
\end{aligned}
\end{eqnarray*}

Similarly, we can get that,
\begin{eqnarray*}
\expected\left[|q'(t)-\hat{q}(t)|\right] \leq 4\hat{\sigma}_{q}(t)(\sqrt{\log TR}+D_{2})
\end{eqnarray*}
where $D_{1}, D_{2}$ are both constant numbers.

Since $\hat{\sigma}_{u_{i}}(t)=\sqrt{\frac{\alpha\hat{u}_{i}(t)(1-\hat{u}_{i}(t))}{T_{i}(t)+1}}+\sqrt{\frac{\beta}{T_{i}(t)}}\leq \frac{\sqrt{\alpha}+\sqrt{\beta}}{\sqrt{T_{i}(t)}}$,
$\hat{\sigma}_{q}(t)=\sqrt{\frac{\alpha\hat{q}(t)(1-\hat{q}(t))}{N_{q}(t)+1}}+\sqrt{\frac{\beta}{N_{q}(t)}}\leq \frac{\sqrt{\alpha}+\sqrt{\beta}}{\sqrt{N_{q}(t)}}$

From (\ref{at1csum}) and Lemma \ref{lm}, we have,
\begin{eqnarray*}
\begin{aligned}
\expected[\Delta U_{t}]
&\leq C_{1}''\expected\left(\sum_{i \in \mathbf{S}^{t}}\frac{\sqrt{\log TR}+D_{1}}{\sqrt{T_{i}(t)}}\right)+C_{2}''\expected\left(\sum_{i \in \mathbf{S}^{t}}(N+1)\frac{\sqrt{\log TR}+D_{2}}{\sqrt{N_{q}(t)}}\right)\\
&+\expected\left[\sum_{i \in \mathbf{S}^{t}}2\big(\sqrt{\frac{\alpha\hat{u}_{i}(t)(1-\hat{u}_{i}(t))\log TR}{T_{i}(t)+1}}+\sqrt{\frac{\beta\log TR}{T_{i}(t)}}\big)\right]\\
&+\expected\left[\sum_{i \in \mathbf{S}^{t}}(N+1)\big(\sqrt{\frac{\alpha\hat{q}(t)(1-\hat{q}(t))\log TR}{N_{q}(t)+1}}+\sqrt{\frac{\beta\log TR}{N_{q}(t)}}\big)\right]\\
&\leq C_{1}'\expected\left(\sum_{i \in \mathbf{S}^{t}}\sqrt{\frac{\log TR}{T_{i}(t)}}\right)+C_{2}'\expected\left(\sum_{i \in \mathbf{S}^{t}}(N+1)\sqrt{\frac{\log TR}{N_{q}(t)}}\right)
\end{aligned}
\end{eqnarray*}
where $C_{1}'$, $C_{2}'$, $C_{1}''$, $C_{2}''$ are absolute constants.

\noindent\textbf{Proof of Lemma \ref{eper30n}:}

\noindent Notation:
\begin{itemize}
    \item
    \begin{eqnarray*}
    \mathbf{S}^{*}\in \arg\max \exputility
    \end{eqnarray*}

    \item
    \begin{eqnarray*}
    \begin{aligned}
    \mathcal{T}&=\{t:u_{i}'(t)\geq u_{i}\textrm{ for all }i\in \mathbf{S}^{*}\},\\
    succ(t)&=\min\{\bar{t}\in \mathcal{T}:\bar{t}>t\},\\
    \varepsilon^{An}(t)&=\{\tau:\tau \in(t,succ(t))\}\textrm{ for all }t\in \mathcal{T}
    \end{aligned}
    \end{eqnarray*}

Here we recall the definition of optimistic round and $\varepsilon^{An}(t)$. $\mathcal{T}$ is the set of ``optimistic" round indices, i.e. when value of $u_{i}'(t)$ is higher than the value of $u_{i}$ for all messages $i$ in the optimal sequential message $\mathbf{S}^{*}$. $succ(t)$ denotes the successive round index after $t$ that is optimistic. $\varepsilon^{An}(t)$ is the set of non-optimistic round between two consecutive optimistic rounds for all $t\in \mathcal{T}$. We will refer to $\varepsilon^{An}(t)$ as the ``analysis round" starting at $t$ round.
    \item 
    \begin{eqnarray*}
    r=\left \lfloor (s+1)^{1/p} \right \rfloor
    \end{eqnarray*}
    \begin{eqnarray*}
    z=\sqrt{\log (rR+1)}
    \end{eqnarray*}

    %and for each messages $i=1,\cdots,N$,
    %\begin{eqnarray}
    %\hat{\sigma}_{u_{i}}(t)=\sqrt{\frac{\alpha\hat{u}_{i}(t)(1-\hat{u}_{i}(t))}{T_{i}(t)+1}}+\sqrt{\frac{\beta}{T_{i}(t)}}
    %\end{eqnarray}

    %\begin{eqnarray}
    %\hat{\sigma}_{q}(t)=\sqrt{\frac{\alpha\hat{q}(t)(1-\hat{q}(t))}{N_{q}(t)+1}}+\sqrt{\frac{\beta}{N_{q}(t)}}
    %\end{eqnarray}
    %where $\alpha$ and $\beta$ are both constant numbers.
    
    \item Define events,
    \begin{eqnarray}
    \label{event}
    \begin{aligned}
        A_{t}&=\bigg\{\{u_{i}'(t)\geq \hat{u}_{i}(t)+z\hat{\sigma}_{u_{i}}(t)\textrm{ for all } i\in \mathbf{S}^{*}\}\cap\{q'(t)\geq \hat{q}(t)+z\hat{\sigma}_{q}(t)\}\bigg\}\\
        B_{t}&=\bigg\{\{\hat{u}_{i}(t)+z\hat{\sigma}_{u_{i}}(t)\geq u_{i}\textrm{ for all }i\in \mathbf{S}^{*}\}\cap\{\hat{q}(t)+z\hat{\sigma}_{q}(t)\geq q\}\bigg\}\\
        \mathfrak{B}_{\tau}&=\bigcap_{t=\tau+1}^{\tau+r}B_{t}
    \end{aligned}
    \end{eqnarray}

\end{itemize}

We have,
\begin{eqnarray*}
P\{|\varepsilon^{An}(\tau)|^{p}<s+1\}=P\{|\varepsilon^{An}(\tau)|\leq r\}
\end{eqnarray*}

By definition, length of the analysis round, $\varepsilon^{An}(\tau)$ less than $r$, implies that one of the rounds from $\tau+1,\cdots,\tau+r$ is optimistic. Hence, we have
\begin{eqnarray*}
\begin{aligned}
P\{|\varepsilon^{An}(\tau)|\leq r\}&=
P\left(\left\{\{u_{i}'\geq u_{i}\textrm{ for all }i\in \mathbf{S}^{*}\}\cap \{q'(t)\geq q\} \textrm{ for some } t\in(\tau,\tau+r]\right\}\right)\\
&\geq P\Bigg(\Big\{\{u_{i}'\geq \hat{u}_{i}(t)+z\hat{\sigma}_{u_{i}}(t) \geq u_{i}\textrm{ for all } i\in \mathbf{S}^{*}\}\\
&\quad\quad \cap \{q'(t)\geq \hat{q}(t)+z\hat{\sigma}_{q}(t)\geq q\}\textrm{ for some } t\in(\tau,\tau+r]\Big\}\Bigg)
\end{aligned}
\end{eqnarray*}

From (\ref{event}), we have,
\begin{eqnarray}
\label{eqetr}
\begin{aligned}
    P\{|\varepsilon^{An}(\tau)|\leq r\}
    &\geq P(\bigcup_{t=\tau+1}^{\tau+r}A_{t}\cap B_{t})\\
    &=1-P(\bigcap_{t=\tau+1}^{\tau+r}A_{t}^{c}\cup B_{t}^{c})
\end{aligned}
\end{eqnarray}

We focus on the term, $P\left(\bigcap_{t=\tau+1}^{\tau+r}A_{t}^{c}\cup B_{t}^{c}\right)$
\begin{eqnarray}
\label{eqpatcbtc}
\begin{aligned}
P\left(\bigcap_{t=\tau+1}^{\tau+r}A_{t}^{c}\cup B_{t}^{c}\right)
&=P\left(\{\bigcap_{t=\tau+1}^{\tau+r}A_{t}^{c}\cup B_{t}^{c}\}\cap \mathfrak{B}_{\tau}\right)+P\left(\{\bigcap_{t=\tau+1}^{\tau+r}A_{t}^{c}\cup B_{t}^{c}\}\cap \mathfrak{B}_{\tau}^{c}\right)\\
&\overset{d_{1}}{\leq} P\left(\bigcap_{t=\tau+1}^{\tau+r}A_{t}^{c}\right)+P(\mathfrak{B}_{\tau}^{c})\\
&\leq P\left(\bigcap_{t=\tau+1}^{\tau+r}A_{t}^{c}\right)+\sum_{t=\tau+1}^{\tau+r}P(B_{t}^{c})
\end{aligned}
\end{eqnarray}
where the inequality follows from union bound. Inequality $(d_{1})$ holds because we observe that
\begin{eqnarray*}
\begin{aligned}
P\left(\big\{\bigcap_{t=\tau+1}^{\tau+r}A_{t}^{c}\cup B_{t}^{c}\big\}\cap \mathfrak{B}_{\tau}\right)
&=P\left(\big\{\bigcap_{t=\tau+1}^{\tau+r}A_{t}^{c}\cup B_{t}^{c}\big\}\cap\big\{ \bigcap_{t=\tau+1}^{\tau+r}B_{t}\big\}\right)\\
&=P\left(\bigcap_{t=\tau+1}^{\tau+r}(A_{t}^{c}\cup B_{t}^{c})\cap B_{t}\right)\\
&=P\left(\bigcap_{t=\tau+1}^{\tau+r}(A_{t}^{c}\cap B_{t})\cup(B_{t}^{c}\cap B_{t})\right)\\
&=P\left(\bigcap_{t=\tau+1}^{\tau+r}(A_{t}^{c}\cap B_{t})\right)\\
&\leq P\left(\bigcap_{t=\tau+1}^{\tau+r}A_{t}^{c}\right) 
\end{aligned}
\end{eqnarray*}

Note that,
\begin{eqnarray}
\label{eqbtc}
\begin{aligned}
P(B_{t}^{c})
&=P\left(\{\bigcup_{i\in \mathbf{S}^{*}}\{\hat{u}_{i}(t)+z\hat{\sigma}_{u_{i}}(t)<u_{i}\}\}\cup \{\hat{q}(t)+z\hat{\sigma}_{q}(t)<q\} \right)\\
&\leq \left(\sum_{i\in \mathbf{S}^{*}}P(\hat{u}_{i}(t)+z\hat{\sigma}_{u_{i}}(t)<u_{i})\right)+P(\hat{q}(t)+z\hat{\sigma}_{q}(t)<q)
\end{aligned}
\end{eqnarray}

Substituting $\rho=rR+1$ in Lemma \ref{lm}, we obtain,
\begin{eqnarray}
\label{eqrk}
\begin{aligned}
P(\hat{u}_{i}(t)+z\hat{\sigma}_{u_{i}}(t)<u_{i})&\leq \frac{1}{(rR+1)^{2\beta}}
\leq \frac{1}{r^{2\beta}R^{2\beta}}\\
P(\hat{q}(t)+z\hat{\sigma}_{q}(t)<q)&\leq \frac{1}{(rR+1)^{2\beta}}
\leq \frac{1}{r^{2\beta}R^{2\beta}}
\end{aligned}
\end{eqnarray}

From (\ref{eqbtc}) and (\ref{eqrk}), we obtain,
\begin{eqnarray}
\label{eqpbtc}
\begin{aligned}
&P(B_{t}^{c})\leq \frac{|\mathbf{S}^{*}|+1}{r^{2\beta}R^{2\beta}}\\
&\sum_{t=\tau+1}^{\tau+r}P(B_{t}^{c})\leq \frac{|\mathbf{S}^{*}|+1}{r^{2\beta-1}R^{2\beta}}\leq \frac{N+1}{r^{2\beta-1}R^{2\beta}}%\leq \frac{1}{r^{3}R^{2}}
\end{aligned}
\end{eqnarray}

We will now use the tail bounds for Gaussian random variables to bound the probability $P(A_{t}^{c})$. For any Gaussian random variable $Z$ with mean $\mu$ and standard deviation $\sigma$, we have,
\begin{eqnarray*}
P(Z>\mu+x\sigma)\geq\frac{1}{\sqrt{2\pi}}\frac{x}{x^{2}+1}e^{-x^{2}/2}
\end{eqnarray*}

Note that by construction of $u_{i}'(t)$ in Algorithm \ref{sbors}, we have,
\begin{eqnarray*}
P\left(\bigcap_{t=\tau+1}^{\tau+r}A_{t}^{c}\right)=P\left(\theta^{(j)}(t)\leq z\textrm{ for all }t\in (\tau,\tau+r]\textrm{ and for all } j=1,\cdots,R\right)
\end{eqnarray*}

Since $\theta^{(j)}(t)$, $j=1,\cdots,R$, $t=\tau+1,\cdots,\tau+r$ are independently sampled from Gaussian distribution $N(0,1)$, we have 
\begin{eqnarray}
\label{eqpatc}
\begin{aligned}
P\left\{\bigcap_{t=\tau+1}^{\tau+r}A_{t}^{c}\right\}
&\leq \left[1-\left(\frac{1}{\sqrt{2\pi}}\frac{\sqrt{\log (rR+1)}}{\log (rR+1)+1}\cdot\frac{1}{\sqrt{rR+1}}\right)\right]^{rR}\\
&\leq \exp\left(-\frac{r^{1/2}}{\sqrt{2\pi}}\frac{2\sqrt{\log (rR+1)}}{4\log (rR+1)+1}\right)\\
&\leq \frac{1}{(rR)^{2.2}}\textrm{ for any }r\geq \frac{e^{12}}{R}
\end{aligned}
\end{eqnarray}

From (\ref{eqetr}), (\ref{eqpatcbtc}), (\ref{eqpbtc}), (\ref{eqpatc}), we have that
\begin{eqnarray*}
P\{|\varepsilon^{An}(\tau)|\leq r\}\geq 1-\frac{N+1}{r^{2\beta-1}R^{2\beta}}-\frac{1}{(rR)^{2.2}}\textrm{ for any } r\geq \frac{e^{12}}{R}
\end{eqnarray*}

From definition $r\geq (s+1)^{1/p}-1$, we obtain
\begin{eqnarray*}
P\{|\varepsilon^{An}(\tau)|<s+1\}\geq 1-\frac{N+1}{(s+1)^{(2\beta-1)/p}-1}-\frac{1}{(s+1)^{2.2/p}-1}\textrm{ for any } s\geq (\frac{e^{12}}{R}+1)^{p}
\end{eqnarray*}

Therefore, we have,
\begin{eqnarray*}
\begin{aligned}
\expected[|\varepsilon^{An}(\tau)|^{p}]
&=\sum_{q=0}^{\infty}P{|\varepsilon^{An}(\tau)|^{p}\geq t}\\
&\leq \left(\frac{e^{12}}{R}+1\right)^{p}+\sum_{s=\frac{e^{12p}}{R^{p}}}^{\infty}P{|\varepsilon^{An}(\tau)|^{p}\geq t}\\
&\leq \left(\frac{e^{12}}{R}+1\right)^{p}+\sum_{s=\frac{e^{12p}}{R^{p}}}^{\infty}\frac{N+1}{s^{(2\beta-1)/p}}+\frac{1}{s^{2.2/p}}\\
&\overset{d_{2}}{\leq} \left(\frac{e^{12}}{R}+1\right)^{p}+C_{3}'N+C_{4}'
\end{aligned}
\end{eqnarray*}
where $C_{3}'$ and $C_{4}'$ are constants. Inequality $(d_{2})$ holds because of Riemann zeta function. Since $\beta\geq 2$ by definition, $(2\beta-1)/p>1$, $2.2/p>1$, the summation of $\frac{1}{s^{(2\beta-1)/p}}$ and $\frac{1}{s^{2.2/p}}$ converge to constants.

The result follows from the above inequality.

\section*{Appendix B: Proofs of Theorems}
\noindent\textbf{Proof of Theorem \ref{theorder}:}
The detail can be seen in \cite{aroyo2018reports} Proof of Theorem 1.

\noindent\textbf{Proof of Theorem \ref{thereg}:}
\begin{eqnarray*}
Reg(T;\mathbf{u},q)
=\expected\left[\sum_{t=1}^{T}\expected[U(\mathbf{S}^{*};\mathbf{u},q)]-\expected[U(\mathbf{S}^{t};\mathbf{u},q)]\right]=Reg_{1}(T;\mathbf{u},q)+Reg_{2}(T;\mathbf{u},q)
\end{eqnarray*}
where $Reg_{1}(T;\mathbf{u},q)=\expected\left[\sum_{t=1}^{T}\expected[U(\mathbf{S}^{*};\mathbf{u},q)]-\expected[U(\mathbf{S}^{t};\mathbf{u}'(t),q'(t))]\right]$,

$Reg_{2}(T;\mathbf{u},q)=\expected\left[\sum_{t=1}^{T}\expected[U(\mathbf{S}^{t};\mathbf{u}'(t),q'(t))]-\expected[U(\mathbf{S}^{t};\mathbf{u},q)]\right]$.

\noindent\textbf{New Notations:}
\begin{itemize}
    \item For any $t,\tau\leq T$, define $\Delta U_{t}$ and $\Delta U_{t,\tau}$ as follows,
    \begin{eqnarray*}
    \Delta U_{t,\tau}\overset{\Delta}{=}\expected[U(\mathbf{S}^{t};\mathbf{u}'(t),q'(t))]-\expected[U(\mathbf{S}^{t};\mathbf{u}^{'}(\tau),q'(\tau))]
    \end{eqnarray*}
\end{itemize}

\noindent\textbf{Old Notations:} (Same as notations in proof of Lemma \ref{disexp} and Lemma \ref{eper30n}.)

\begin{itemize}
    \item  For any $t$, define $\Delta U_{t}$ as follows,
    \begin{eqnarray*}
    \Delta U_{t}\overset{\Delta}{=}\expected[U(\mathbf{S}^{t};\mathbf{u}'(t),q'(t))]-\expected[U(\mathbf{S}^{t};\mathbf{u},q)]
    \end{eqnarray*}
    
    \item For any $t\in \{1,\cdots,T\}$, define events $\mathcal{G}_{t}$, $\mathcal{H}_{t}$ as
    \begin{eqnarray*}
    \mathcal{G}_{t}=\left\{|\hat{u}_{i}(t)-u_{i}|\geq \sqrt{\frac{\alpha\hat{u}_{i}(t)(1-\hat{u}_{i}(t))\log t}{T_{i}(t)+1}}+\sqrt{\frac{\beta\log t}{T_{i}(t)}}\textrm{ for some } i=1,\cdots,N\right\}
    \end{eqnarray*}
    \begin{eqnarray*}
    \mathcal{H}_{t}=\left\{|\hat{q}(t)-q|\geq \sqrt{\frac{\alpha\hat{q}(t)(1-\hat{q}(t))\log t}{N_{q}(t)+1}}+\sqrt{\frac{\beta\log t}{N_{q}(t)}}\right\}
    \end{eqnarray*}
    
    where the definition of $\hat{\sigma}_{u_{i}}(t)$ and $\hat{\sigma}_{q}(t)$ can be seen in Algorithm \ref{sbors}.
    
    \item
    Define events $\mathcal{A}_{t}=\mathcal{G}_{t}\cap \mathcal{H}_{t}$
    
    \item
    \begin{eqnarray*}
    \mathbf{S}^{*}\in \arg\max\expected[U(\mathbf{S};\mathbf{u},q)]
    \end{eqnarray*}
    
    \item
    \begin{eqnarray*}
    \begin{aligned}
    \mathcal{T}&=\{t:u_{i}'(t)\geq u_{i}\textrm{ for all }i\in \mathbf{S}^{*}\},\\
    succ(t)&=\min\{\bar{t}\in \mathcal{T}:\bar{t}>t\},\\
    \varepsilon^{An}(t)&=\{\tau:\tau \in(t,succ(t))\}\textrm{ for all }t\in \mathcal{T}
    \end{aligned}
    \end{eqnarray*}
\end{itemize}

\textbf{Bounding $Reg_{2}(T,\mathbf{u},q)$: }

Note that $Reg_{2}(T,\mathbf{u},q)=\expected\left[\sum_{t=1}^{T}\big(\expected[U(\mathbf{S}^{t};\mathbf{u}'(t),q'(t))]-\expected[U(\mathbf{S}^{t};\mathbf{u},q)]\big)\right]=\expected\left\{\sum_{t=1}^{T}\Delta U_{t}\right\}$

From Lemma \ref{disexp}, we have,
\begin{eqnarray*}
\begin{aligned}
Reg_{2}(T,\mathbf{u},q)
&\leq C_{1}'\expected\left(\sum_{t=1}^{T}\sum_{i \in \mathbf{S}^{t}}\sqrt{\frac{\log TR}{T_{i}(t)}}\right)+C_{2}'\expected\left(\sum_{t=1}^{T}\sum_{i \in \mathbf{S}^{t}}(N+1)\sqrt{\frac{\log TR}{N_{q}(t)}}\right)
\end{aligned}
\end{eqnarray*}
where $C_{1}'$, $C_{2}'$ are absolute constants.

Denote $n_{i}$ as the total number of rounds that message $i$ is in the sequence, then we have,
\begin{eqnarray*}
\sum_{T_{i}(t)=1}^{n_{i}}\frac{1}{\sqrt{T_{i}(t)}}\leq 2\sqrt{n_{i}}
%\sum_{N_{q}(t)=1}^{n_{i}}\frac{1}{\sqrt{N_{q}(t)}}\leq 2\sqrt{n_{i}}
\end{eqnarray*}

Thus,
\begin{eqnarray*}
\begin{aligned}
\sum_{t=1}^{T}\sum_{i \in \mathbf{S}^{t}}\sqrt{\frac{\log TR}{T_{i}(t)}}
&=\sum_{i=1}^{N}\sum_{T_{i}(t)=1}^{n_{i}}\sqrt{\frac{\log TR}{T_{i}(t)}}\\
&\leq 2\sum_{i=1}^{N}\sqrt{T\log TR}\\
&\leq 2N\sqrt{T\log TR}
\end{aligned}
\end{eqnarray*}

Similarly, we can get
\begin{eqnarray*}
\sum_{t=1}^{T}\sum_{i \in \mathbf{S}^{t}}(N+1)\sqrt{\frac{\log TR}{N_{q}(t)}}
\leq 2(N+1)N\sqrt{T\log TR}
\end{eqnarray*}

As a result,
\begin{eqnarray*}
\begin{aligned}
Reg_{2}(T,\mathbf{u},q)
&\leq 2C_{1}'N\sqrt{T\log TR}+2C_{2}'N(N+1)\sqrt{T\log TR}\\
&\leq C_{1}N^{2}\sqrt{NT\log TR}
\end{aligned}
\end{eqnarray*}
where $C_{1}$ is a constant number.

\textbf{Bounding $Reg_{1}(T,\mathbf{u},q)$: }

Recall that $\mathcal{T}$ is the set of optimistic round and the analysis epoch $\varepsilon^{An}(t)$ is the set of non-optimistic rounds between optimistic round $t$ and its subsequent optimistic round. Therefore, we can reformulate $Reg_{1}(T,\mathbf{u},q)$ as,
\begin{eqnarray*}
\begin{aligned}
Reg_{1}(T,\mathbf{u},q)
&=\expected\left[\sum_{t=1}^{T}\expected[U(\mathbf{S}^{*};\mathbf{u},q)]-\expected[U(\mathbf{S}^{t};\mathbf{u}'(t),q'(t))]\right]\\
&\overset{e_{1}}{\leq}\expected\left[\sum_{t=1}^{T}\mathbbm{1}(t \in \mathcal{T})\cdot \sum_{\tau \in \epsilon^{An}(t)}\expected[U(\mathbf{S}^{*};\mathbf{u},q)]-\expected[U(\mathbf{S}^{\tau};\mathbf{u}^{'}(\tau),q'(\tau))]\right]\\
&\overset{e_{2}}{\leq}
\expected\left[\sum_{t=1}^{T}\mathbbm{1}(t \in \mathcal{T})\cdot \sum_{\tau \in \epsilon^{An}(t)}\expected[U(\mathbf{S}^{t};\mathbf{u}'(t),q'(t))]-\expected[U(\mathbf{S}^{\tau};\mathbf{u}^{'}(\tau),q'(\tau))]\right]\\
&\overset{e_{3}}{\leq}
\expected\left[\sum_{t=1}^{T}\mathbbm{1}(t \in \mathcal{T})\cdot \sum_{\tau \in \epsilon^{An}(t)}\expected[U(\mathbf{S}^{t};\mathbf{u}'(t),q'(t))]-\expected[U(\mathbf{S}^{t};\mathbf{u}^{'}(\tau),q'(\tau))]\right]\\
&=\expected\left[\sum_{t=1}^{T}\mathbbm{1}(t \in \mathcal{T})\cdot \sum_{\tau \in \epsilon^{An}(t)}\Delta U_{t,\tau}\right]
\end{aligned}
\end{eqnarray*}

Inequality $(e_{1})$ and $e_{2}$ hold because $\mathbf{S}^{t}$ is the optimal message sequence when parameters are given by $\mathbf{u}(t), q(t)$ such that $\expected[U(\mathbf{S}^{t};\mathbf{u}'(t),q'(t))]\geq \expected[U(\mathbf{S}^{*};\mathbf{u}'(t),q'(t))]$ for any $t$. The restricted monotonicity property in Lemma \ref{lip} implies $\expected[U(\mathbf{S}^{*};\mathbf{u}'(t),q'(t))]\geq \expected[U(\mathbf{S}^{*};\mathbf{u},q)]$ for any $t\in \mathcal{T}$. Therefore we can drop the optimistic rounds.

Inequality $(e_{3})$ follows from the observation that by design for any $\tau$, $\expected[U(\mathbf{S}^{\tau};u^{'}(\tau),q'(\tau))]\geq \expected[U(\mathbf{S};u^{'}(\tau),q'(\tau))]$ for any sequential messages $S$. Therefore,  $\expected[U(\mathbf{S}^{\tau};u^{'}(\tau),q'(\tau))]\geq \expected[U(\mathbf{S}^{t};u^{'}(\tau),q'(\tau))]$ holds for any $\tau, t$.

Following the approach of proving Lemma \ref{disexp}, we analyze the first term, $Reg_{1}(T,\mathbf{u},q)$ in two scenarios, one when $\mathcal{A}_{t}\cup \mathcal{A}_{\tau}$ is true and another when $(\mathcal{A}_{t}\cup \mathcal{A}_{\tau})^{c}$ is true. More specifically, 
\begin{eqnarray*}
\expected\left[\sum_{\tau \in \epsilon^{An}(t)}\Delta U_{t,\tau}\right]=
\expected\left[\sum_{\tau \in \epsilon^{An}(t)}\Delta U_{t,\tau}\cdot \mathbbm{1}(\mathcal{A}_{t}\cup \mathcal{A}_{\tau})+\Delta U_{t,\tau}\cdot \mathbbm{1}((\mathcal{A}_{t}\cup \mathcal{A}_{\tau})^{c})\right]
\end{eqnarray*}

From the Lipschitz property in Lemma \ref{lip} and triangle inequality, we obtain
\begin{eqnarray*}
\begin{aligned}
\Delta U_{t,\tau}
&=\expected[\mathbf{S}^{t};u^{'}(t),q'(t)]-\expected[\mathbf{S}^{t};u^{'}(\tau),q'(\tau)]\\
&\leq 
\sum_{i \in \mathbf{S}^{t}}\left(2|u_{i}'(t)-u_{i}|+(N+1)|q'(t)-q|\right)+
\sum_{i \in \mathbf{S}^{t}}\left(2|u_{i}'(\tau)-u_{i}|+(N+1)|q'(\tau)-q|\right)
\end{aligned}
\end{eqnarray*}

We have,
\begin{eqnarray*}
\begin{aligned}
\expected\left[\sum_{\tau \in \varepsilon^{An}(t)}\Delta U_{t,\tau}\right]
&\leq \expected\bigg[N\cdot|\varepsilon^{An}(t)|\cdot \mathbbm{1}(\mathcal{A}_{t}\cup \mathcal{A}_{\tau})
+\mathbbm{1}((\mathcal{A}_{t}\cup \mathcal{A}_{\tau})^{c})\cdot \sum_{\tau\in \varepsilon^{An}(t)}\bigg(\sum_{i \in \mathbf{S}^{t}}\big(2|u_{i}'(t)-u_{i}|\\
&+(N+1)|q'(t)-q|\big)+\sum_{i \in \mathbf{S}^{t}}\big(2|u_{i}'(\tau)-u_{i}|+(N+1)|q'(\tau)-q|\big)\bigg)\bigg]
\end{aligned}
\end{eqnarray*}

Similar to the proof of Lemma \ref{disexp}, we can show that
\begin{eqnarray*}
\begin{aligned}
&\expected\bigg[\mathbbm{1}((\mathcal{A}_{t}\cup \mathcal{A}_{\tau})^{c})\cdot \sum_{\tau\in \varepsilon^{An}(t)}\bigg(\sum_{i \in \mathbf{S}^{t}}\big(2|u_{i}'(t)-u_{i}|+(N+1)|q'(t)-q|\big)\\
&\quad\quad+\sum_{i \in \mathbf{S}^{t}}\big(2|u_{i}'(\tau)-u_{i}|+(N+1)|q'(\tau)-q|\big)\bigg)\bigg]\\
&\leq \expected\bigg[\sum_{\tau\in \varepsilon^{An}(t)}\bigg(\mathbbm{1}(\mathcal{A}_{t}^{c})\cdot\sum_{i \in \mathbf{S}^{t}}\big(2|u_{i}'(t)-u_{i}|+(N+1)|q'(t)-q|\big)\\
&\quad\quad+\mathbbm{1}(\mathcal{A}_{\tau}^{c})\cdot\sum_{i \in \mathbf{S}^{t}}\big(2|u_{i}'(\tau)-u_{i}|+(N+1)|q'(\tau)-q|\big)\bigg)\bigg]\\
&\leq \expected\left(|\varepsilon^{An}(t)|\cdot \sum_{i \in \mathbf{S}^{t}}\big(C_{1}'\sqrt{\frac{\log TR}{T_{i}(t)}}+C_{2}'(N+1)\sqrt{\frac{\log TR}{N_{q}(t)}}\big)\right)
\end{aligned}
\end{eqnarray*}
where $C_{1}'$ and $C_{2}'$ are constant numbers. As a result, we have that
\begin{eqnarray}
\label{reg1sep}
\begin{aligned}
&Reg_{1}(T,\mathbf{u},q)\\
&\leq \expected\left[\sum_{t=1}^{T}\bigg(N\cdot|\varepsilon^{An}(t)|\cdot \mathbbm{1}(\mathcal{A}_{t})+|\varepsilon^{An}(t)|\cdot\sum_{i \in \mathbf{S}^{t}}\big(C_{1}'\sqrt{\frac{\log TR}{T_{i}(t)}}+C_{2}'(N+1)\sqrt{\frac{\log TR}{N_{q}(t)}}\big)\bigg)\right]
\end{aligned}
\end{eqnarray}

We bound each of term in the above expression to complete the proof. We have by Cauchy-Schwartz inequality.
\begin{eqnarray*}
\begin{aligned}
\expected\big[|\varepsilon^{An}(t)|\cdot \mathbbm{1}(\mathcal{A}_{t})\big]
\leq \expected^{1/2}\big(|\varepsilon^{An}(t)|^{2}\big)\cdot P^{1/2}(\mathcal{A}_{t})
\end{aligned}
\end{eqnarray*}

Since in Lemma \ref{eper30n}, we show that
$\expected^{1/2}[|\varepsilon^{An}(\tau)|^{2}]\leq \frac{e^{12}}{R}+(C_{3}'N)^{1/2}+C_{4}'^{1/2}$. Based on Lemma \ref{lm}, we obtain that $P(\mathcal{A}_{t})\leq \frac{2N}{t^{2\beta}}\times \frac{2}{t^{2\beta}}=\frac{4N}{t^{4\beta}}$
Therefore, we have
\begin{eqnarray}
\label{reg1e13}
\expected\left[\sum_{t=1}^{T}N\cdot|\varepsilon^{An}(t)|\cdot \mathbbm{1}(\mathcal{A}_{t})\right]
\leq \sum_{t=1}^{T}N\left(\frac{e^{12}}{R}+(C_{3}'N)^{1/2}+C_{4}^{1/2}\right)\cdot \frac{2}{t^{4\beta}}
\leq N\big(\frac{e^{13}}{R}+C_{3}''\sqrt{N}+C_{4}''\big)
\end{eqnarray}
where $C_{3}'$. $C_{4}'$, $C_{3}''$, $C_{4}''$ are constants.

Now we bound the second term in (\ref{reg1sep}). We make the following notation for brevity.
\begin{eqnarray*}
\delta_{i}(t)=C_{1}'\sum_{t\in \mathbf{S}^{t}}\sqrt{\frac{\log TR}{T_{i}(t)}}\\
\Delta_{i}(t)=C_{2}'\sum_{t\in \mathbf{S}^{t}}(N+1)\sqrt{\frac{\log TR}{N_{q}(t)}}
\end{eqnarray*}

From Cauchy-Schwartz inequality, we have
\begin{eqnarray*}
\sum_{t=1}^{T}|\varepsilon^{An}(t)|\big(\delta_{i}(t)+\Delta_{i}(t)\big)\leq \left(\sum_{t=1}^{T}|\varepsilon^{An}(t)|^{2}\right)^{1/2}\cdot
\left[\bigg(\sum_{t=1}^{T}\delta_{i}^{2}(t)\bigg)^{1/2}+\bigg(\sum_{t=1}^{T}\Delta_{i}^{2}(t)\bigg)^{1/2}\right]
\end{eqnarray*}

Again applying Cauchy-Schwartz on $\delta_{i}(t)$ and $\Delta_{i}(t)$, we have
\begin{eqnarray*}
\delta^{2}_{i}(t)\leq C_{1}^{'2}\left(\sum_{i\in \mathbf{S}^{t}}1\cdot \sum_{i\in \mathbf{S}^{t}}\frac{\log TR}{T_{i}(t)}\right)
\leq C_{1}^{'2}N\cdot \sum_{i\in \mathbf{S}^{t}}\frac{\log TR}{T_{i}(t)}
\\
\Delta^{2}_{i}(t)\leq C_{2}^{'2}\left(\sum_{i\in \mathbf{S}^{t}}1\cdot \sum_{i\in \mathbf{S}^{t}}\frac{\log TR}{N_{q}(t)}\right)
\leq C_{2}^{'2}N\cdot \sum_{i\in \mathbf{S}^{t}}\frac{\log TR}{N_{q}(t)}
\end{eqnarray*}

Recall that $n_{i}$ denote the total number of rounds that message $i$ is in the sequence, thus we have
\begin{eqnarray*}
\sum_{t=1}^{T}\sum_{i\in \mathbf{S}^{t}}\frac{\log TR}{T_{i}}
=\sum_{i=1}^{N}\sum_{T_{i}(t)=1}^{n_{i}}\frac{\log TR}{T_{i}(t)}
\overset{f_{1}}{\leq} N\log TR\cdot \log T\\
\sum_{t=1}^{T}\sum_{i\in \mathbf{S}^{t}}\frac{\log TR}{N_{q}}
=\sum_{i=1}^{N}\sum_{N_{q}(t)=1}^{n_{i}}\frac{\log TR}{N_{q}(t)}
\overset{f_{2}}{\leq} N\log TR\cdot \log T
\end{eqnarray*}
Inequality $(f_{1}), (f_{2})$ hold because $\sum_{T_{i}(t)}^{n_{i}}\frac{1}{T_{i}(t)}\leq \log n_{i}$,
$\sum_{N_{q}(t)}^{n_{i}}\frac{1}{N_{q}(t)}\leq \log n_{i}$.

Due to Jensen's inequality and Lemma \ref{eper30n} (substitute p=1), we have
\begin{eqnarray*}
\begin{aligned}
\expected\left[\left(\sum_{t=1}^{T}|\varepsilon^{An}(t)|^{2}\right)^{1/2}\right]
\leq \left(\expected\left[\sum_{t=1}^{T}|\varepsilon^{An}(t)|\right]\right)^{1/2}=\left(\expected\left[\sum_{t=1}^{T}\frac{e^{12}}{R}+C_{3}'N+C_{4}'\right]\right)^{1/2}\leq C_{5}'\sqrt{NT}
\end{aligned}
\end{eqnarray*}
where $C_{3}'$, $C_{4}'$ and $C_{5}'$ are constants.

According to (\ref{reg1e13}), we have that,
\begin{eqnarray*}
\expected\left[\left(\sum_{t=1}^{T}|\varepsilon^{An}(t)|^{2}\right)^{1/2}\cdot\bigg(\sum_{t=1}^{T}\delta_{i}^{2}(t)\bigg)^{1/2}\right]
\leq C_{3}''N\sqrt{T\log TR\cdot \log T}\\
\expected\left[\left(\sum_{t=1}^{T}|\varepsilon^{An}(t)|^{2}\right)^{1/2}\cdot\bigg(\sum_{t=1}^{T}\Delta_{i}^{2}(t)\bigg)^{1/2}\right]
\leq C_{4}''N\sqrt{T\log TR\cdot \log T}
\end{eqnarray*}

Hence, from the preceding two results, we have
\begin{eqnarray*}
Reg_{1}(T;\mathbf{u},q)\leq C_{2}N\sqrt{T\log TR\cdot \log T}+\frac{C_{3}N}{R}
\end{eqnarray*}
where $C_{2}, C_{3}, C_{3}'', C_{4}''$ is constant numbers.

As a result, 
\begin{eqnarray*}
\begin{aligned}
Reg(T;\mathbf{u},q)
&=Reg_{2}(T;\mathbf{u},q)+Reg_{1}(T;\mathbf{u},q)\\
&\leq C_{1}N^{2}\sqrt{NT\log TR}+C_{2}N\sqrt{T\log TR\cdot \log T}+\frac{C_{3}N}{R}.
\end{aligned}
\end{eqnarray*}

\end{document}